\documentclass[10pt,twocolumn,letterpaper]{article}
\usepackage[T1]{fontenc}
\usepackage{fix-cm} 

\usepackage[pagenumbers]{cvpr}

\def\arxivbuild{} 

\ifdefined\arxivbuild\else
  \usepackage{cuted} 
\fi

\usepackage{currfile} 

\usepackage{caption} 
\usepackage{placeins}
\usepackage{needspace}
\makeatletter
\renewcommand*{\LenToUnit}[1]{\strip@pt\dimexpr#1\relax}
\makeatother

\newif\ifshowrevisions
\showrevisionstrue
\newcommand{\revisioncolor}{%
  \ifshowrevisions\color{red}\hypersetup{allcolors=red}\fi}
\DeclareRobustCommand{\red}[1]{%
  \ifshowrevisions\textcolor{red}{\hypersetup{allcolors=red}#1}\else#1\fi}
\newenvironment{revisionblock}{\revisioncolor}{\par}
\newcommand{\revisionfloat}{%
  \revisioncolor
  \ifshowrevisions
    \captionsetup{labelfont+={color=red},textfont+={color=red}}%
  \fi}
\AtBeginDocument{\pdfstringdefDisableCommands{\def\red#1{#1}}}
\newsavebox{\revisiongraphicbox}
\newcommand{\revisiongraphic}[2][]{%
  \ifshowrevisions
    \begingroup
    \sbox{\revisiongraphicbox}{\includegraphics[#1]{#2}}%
    \setlength{\fboxsep}{0pt}\setlength{\fboxrule}{0.5pt}%
    \makebox[\wd\revisiongraphicbox][c]{%
      \raisebox{0pt}[\ht\revisiongraphicbox][\dp\revisiongraphicbox]{%
        \fcolorbox{red}{white}{\usebox{\revisiongraphicbox}}}}%
    \endgroup
  \else
    \includegraphics[#1]{#2}%
  \fi}

\newif\ifanonymous
\anonymoustrue
\DeclareRobustCommand{\anon}[2]{\ifanonymous#1\else#2\fi}

\usepackage{microtype}
\ifdefined\pdfinclusioncopyfonts
\fi
\showrevisionsfalse 
\anonymousfalse 
\usepackage{booktabs}
\usepackage{makecell}
\usepackage{algorithm}
\usepackage{algpseudocode}

\definecolor{cvprblue}{rgb}{0.21,0.49,0.74}
\usepackage[pagebackref,breaklinks,colorlinks,allcolors=cvprblue]{hyperref}
\makeatletter
\providecommand*{\theHALG@line}{}
\renewcommand*{\theHALG@line}{\theHalgorithm.\arabic{ALG@line}}
\makeatother

\def\confName{CVPR}
\def\confYear{2027}

\title{From Game Generation to Game Playing:\\Fully Automated Ability Training for Vision-Language-Action Models}

\author{Changran Liu\textsuperscript{1} \quad Qi Wu\textsuperscript{1} \quad Wenchao Ma\textsuperscript{2} \quad
  Sharon X. Huang\textsuperscript{2} \quad Haomiao Jiang\textsuperscript{1}\\
  \textsuperscript{1}Roblox \qquad \textsuperscript{2}The Pennsylvania State University\\[4pt]
  {\small Project page: \url{https://assettool-labs.github.io/script2play/}}}
\hypersetup{pdfauthor={Changran Liu, Qi Wu, Wenchao Ma, Sharon X. Huang, Haomiao Jiang},
  pdftitle={From Game Generation to Game Playing: Fully Automated Ability Training for Vision-Language-Action Models}}

\begin{document}
\maketitle
\setlength{\textfloatsep}{8pt plus 2pt minus 1pt}
\setlength{\dbltextfloatsep}{8pt plus 2pt minus 1pt}
\setlength{\floatsep}{8pt plus 2pt minus 1pt}
\setlength{\dblfloatsep}{8pt plus 2pt minus 1pt}
\captionsetup{skip=3pt}
\begin{abstract}
Training vision-language-action (VLA) models to follow gameplay instructions across diverse games is limited by the cost of collecting demonstrations. We introduce a framework that automatically generates game environments and expert demonstrations. An LLM-powered agent uses game-building tools to construct executable environments and tasks targeting specified skills. Programmatic experts then use privileged simulator state to generate aligned instructions, rendered observations, and keyboard and mouse actions. We train our game policy, \textup{\textsc{Script2Play}}, by post-training a 4B vision-language model with behavior cloning on 150.3 hours of synthetic \red{gameplay in Minecraft and \anon{an in-house game engine}{Roblox}}, without human gameplay demonstrations or reinforcement learning. Evaluation on multiple public game benchmarks demonstrates that \textup{\textsc{Script2Play}} follows gameplay instructions effectively and generalizes across game engines, highlighting the effectiveness of targeted synthetic supervision.

\end{abstract}

\section{Introduction}
\label{sec:intro}

Developing generalist agents that ground free-form language instructions in visual observations and execute low-level actions in interactive environments is a central step toward general-purpose embodied intelligence~\cite{sima2024scaling,simateam2025sima2,wang2025gametars,tan2025lumine}.
Video games offer a rich testbed for studying these capabilities: a single agent must map an instruction and rendered frames into human-compatible actions, such as key presses and mouse movements, across worlds with varying visuals, physics, and interaction logic~\cite{sima2024scaling,guo2026gamewam,zhang2026scaling}.
However, training vision-language-action (VLA) models that generalize across games faces a persistent supervision bottleneck.
Existing approaches largely rely on human gameplay demonstrations~\cite{baker2022vpt,sima2024scaling,li2025jarvisvla,wang2025gametars,yue2025pixels2play,yue2026openp2p} or online reinforcement learning (RL)~\cite{kempka2016vizdoom,ouyang2026player1}.
Collecting human demonstrations with aligned instructions and actions is costly across diverse tasks and titles.
Public gameplay recordings and internet video corpora~\cite{guss2019minerl,baker2022vpt,fan2022minedojo,yue2025pixels2play,choi2026d2e,magne2026nitrogen,yue2026openp2p,markov2026gaming500} may lack explicit task boundaries and verifiable completion signals, or require retrospective captioning.
RL requires interactive access, reward shaping, and extensive rollouts for each game.
Hand-built training suites~\cite{cobbe2020procgen,ouyang2026player1} also require substantial engineering effort whenever a new gameplay skill or mechanic is introduced.

\begin{figure*}[t]
  \centering
    \revisiongraphic[width=0.90\linewidth]{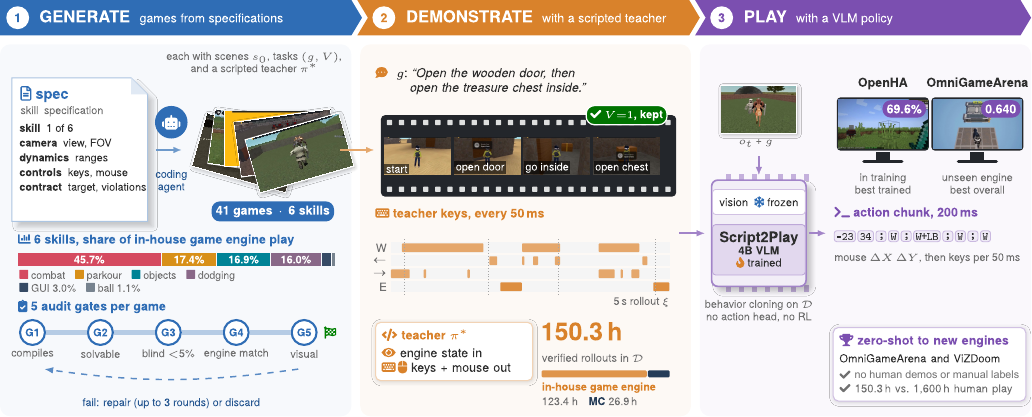}
  \caption{\textbf{Overview.} \emph{Generate:} a coding agent turns each specification into a game with scenes $s_0$, tasks $(g, V)$, and a scripted teacher $\pi^*$; a game must pass five audit gates, and a failing game is repaired (up to three rounds) or discarded.
    \emph{Demonstrate:} the teacher plans on privileged engine state but acts only through keyboard and mouse; the piano roll shows its recorded keys (W: forward, $\leftarrow$: camera left, $\rightarrow$: camera right, E: object interaction) in one rollout, which the verifier accepts ($V = 1$). Only verified rollouts enter $\mathcal{D}$.
    \emph{Play:} behavior cloning on $\mathcal{D}$ turns a pretrained VLM into the policy $\pi_\theta$, which writes each 200\,ms action chunk as text. It is evaluated on OpenHA in Minecraft, a training engine, and zero-shot on OmniGameArena (UE5) and ViZDoom.}
  \label{fig:pipeline}
\end{figure*}

Our core insight is that both skill-targeted game creation and expert demonstration generation can be framed as coding-agent tasks~\cite{aggarwal2026gymanythingrobotics,aggarwal2026gymanythingrt}.
Because game engines represent scenes, mechanics, and task logic as code, an LLM-powered agent can invoke game-building tools from a natural-language skill specification (such as navigation, dynamic obstacle avoidance, moving-target aiming, combat, or object interaction) to construct playable environments paired with programmatic task verifiers.
Crucially, unlike commercial games that expose only rendered pixels, a generated environment exposes its full privileged ground-truth state, including character poses, target locations, enemy trajectories, and collision geometry.
The coding agent can therefore also write programmatic expert controllers (using pathfinding, analytic target tracking, or finite-state machines) that read this privileged state to solve each task while issuing actions strictly through standard keyboard and mouse inputs.
Automated playability and verification checks retain only successful rollouts, turning privileged simulator state into a purely synthetic, data-efficient stream of aligned instructions, frames, and actions.
Every trajectory targets a specified skill, with exact instruction boundaries and verified completion, avoiding the idle traversal and label noise of raw human gameplay.

We obtain \textsc{Script2Play} by post-training Qwen3.5-4B on 150.3 hours of synthetic gameplay: 26.9 hours from Minecraft and 123.4 hours from \red{\anon{an in-house game engine}{Roblox}}, with zero human gameplay demonstrations.
The same unified checkpoint is evaluated on all three benchmarks~\cite{wang2025openha,lin2026omnigamearena,kempka2016vizdoom}.
\textsc{Script2Play} outperforms policies trained on thousands of hours of human play on OpenHA and OmniGameArena~\cite{wang2025gametars,magne2026nitrogen,yue2026openp2p,zhang2026scaling}.
A separate comparison holds the Qwen2.5-VL-7B backbone fixed and uses only our Minecraft subset to compare programmatic supervision with human-play and RL-based training pipelines. These results show that our framework can train game agents with strong instruction-following abilities and cross-engine generalization using a modest amount of scripted demonstration data.

Our contributions are three-fold:
\begin{enumerate}
  \item We design an automated, skill-conditioned game generation framework in which a tool-using coding agent constructs executable game environments, diverse instruction-following tasks, and state-based verifiers directly from natural-language specifications.
  \item We introduce a purely synthetic, data-efficient programmatic expert supervision pipeline that leverages privileged simulator state to synthesize verified, human-compatible keyboard-and-mouse demonstrations for post-training pixel-to-action VLA models without human gameplay data.
  \item Using only 150.3\,h of scripted teacher data, \textsc{Script2Play} outperforms policies trained on thousands of hours of human gameplay on public game benchmarks.
\end{enumerate}

\setcounter{dbltopnumber}{2}
\begin{figure*}[t]
  \revisionfloat
  \centering
  \revisiongraphic[width=0.80\linewidth]{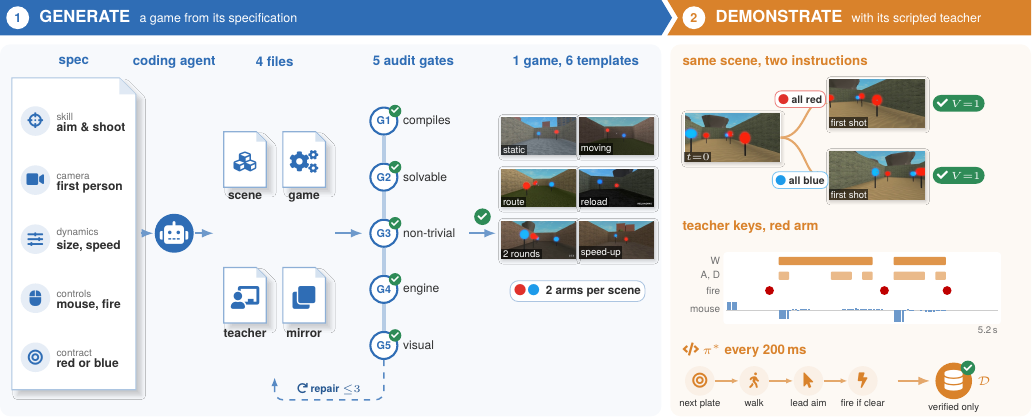}
  \caption{\textbf{Generating a game and its demonstrations.} A coding agent generates a game and scripted teacher $\pi^*$ from a skill specification. Here, two tasks use the same scene but ask the teacher to shoot all red or all blue targets. Only rollouts that complete the instructed task and pass verification enter $\mathcal{D}$.}
  \label{fig:generation}
\end{figure*}

\begin{figure*}[t]
  \revisionfloat
  \centering
  \revisiongraphic[width=0.95\linewidth]{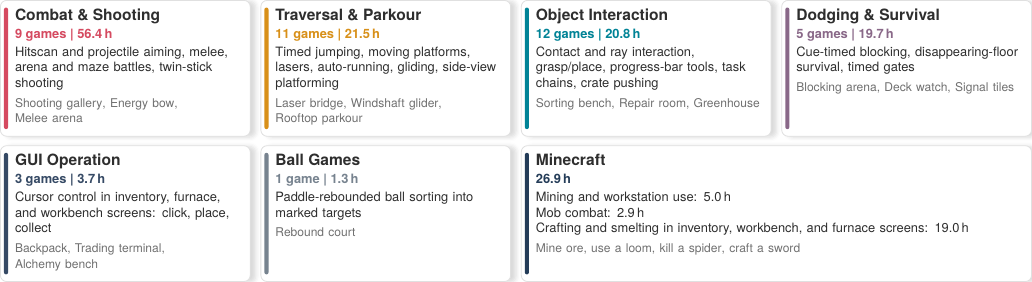}
  \par\vspace{2pt}
  \revisiongraphic[width=0.95\linewidth]{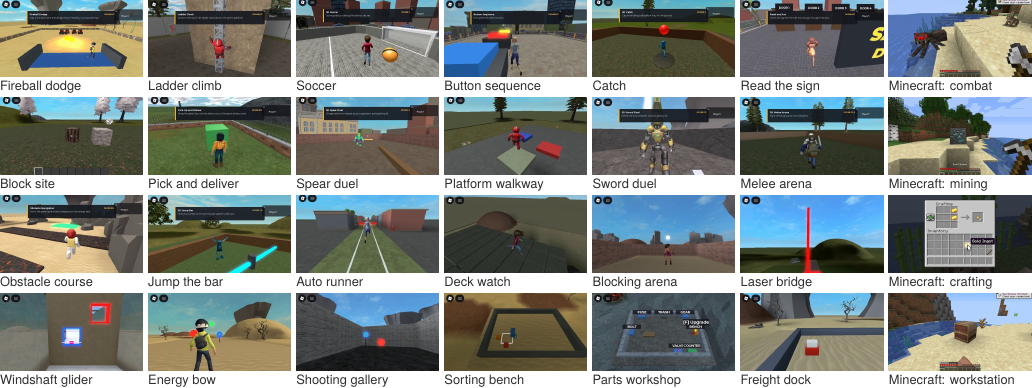}
  \caption{\textbf{Training environments organized by reusable gameplay skills.} Top: skill categories, mechanics, and verified gameplay hours. Bottom: representative frames. \anon{In-house}{Roblox} games cover six categories and Minecraft covers OpenHA's three groups, with multiple camera views.}
  \label{fig:data3d:stats}
  \label{fig:data3d:mosaic}
  \label{fig:data3d:frames}
\end{figure*}

\begin{figure*}[t]
  \centering
  \setlength{\tabcolsep}{1.5pt}
  \resizebox{0.72\linewidth}{!}{\begin{minipage}{\linewidth}\begin{tabular}{@{}*{4}{>{\raggedright\arraybackslash}p{0.245\linewidth}}@{}}
    \includegraphics[width=\linewidth]{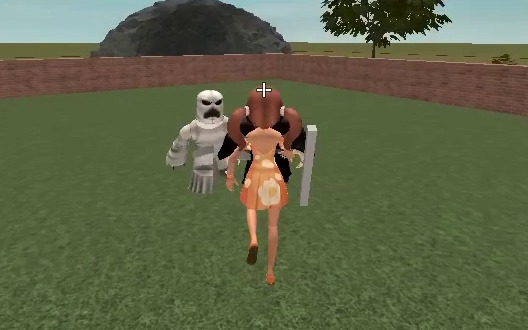}\newline{\scriptsize\textbf{Park}, Belle the Seaside Girl, FOV 67$^\circ$\par}{\tiny\itshape ``Fight off the group of sword fighters.''\par} &
    \includegraphics[width=\linewidth]{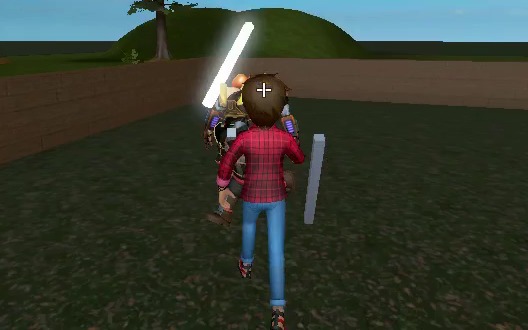}\newline{\scriptsize\textbf{Forest}, Stylish Duke, FOV 75$^\circ$\par}{\tiny\itshape ``Several swordsmen attack you: hit them and stay out of their swings.''\par} &
    \includegraphics[width=\linewidth]{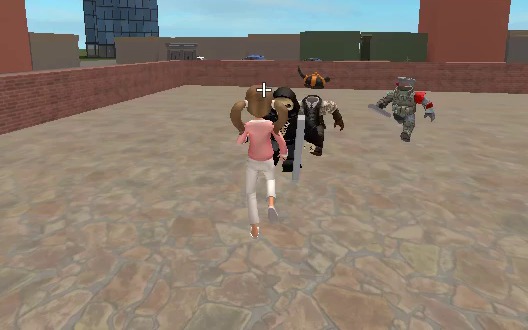}\newline{\scriptsize\textbf{City}, Bonnie on Vacation, FOV 66$^\circ$\par}{\tiny\itshape ``Survive the melee: strike the enemies and dodge their blows.''\par} &
    \includegraphics[width=\linewidth]{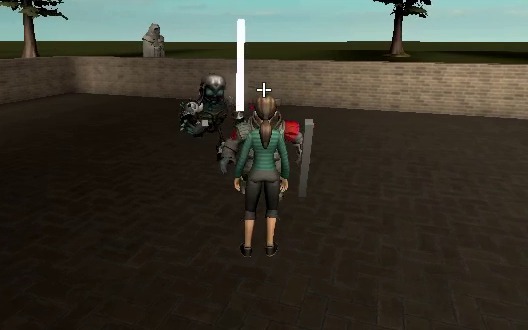}\newline{\scriptsize\textbf{Courtyard}, Summer, FOV 71$^\circ$\par}{\tiny\itshape ``Several swordsmen attack you: hit them and stay out of their swings.''\par} \\[3pt]
    \includegraphics[width=\linewidth]{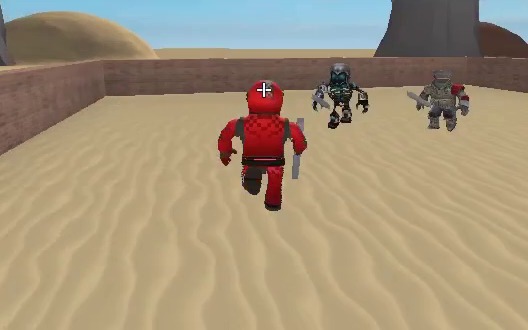}\newline{\scriptsize\textbf{Desert}, \red{\anon{Racer}{ROBLOX Racer}}, FOV 66$^\circ$\par}{\tiny\itshape ``Fight off the group of sword fighters.''\par} &
    \includegraphics[width=\linewidth]{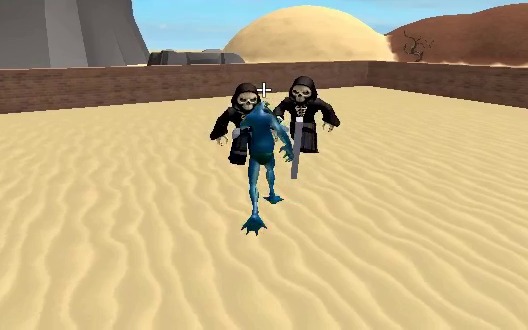}\newline{\scriptsize\textbf{Desert}, Fisherman, FOV 74$^\circ$\par}{\tiny\itshape ``Defeat the enemies around you with your sword and avoid their attacks.''\par} &
    \includegraphics[width=\linewidth]{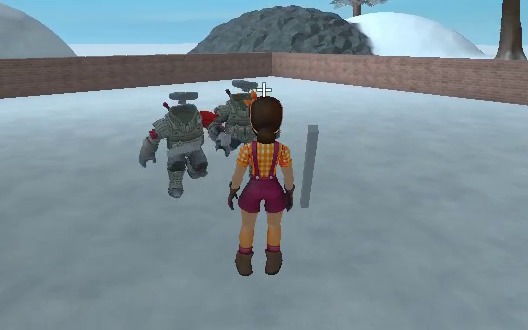}\newline{\scriptsize\textbf{Snow}, Lumber Jessie, FOV 68$^\circ$\par}{\tiny\itshape ``Survive the melee: strike the enemies and dodge their blows.''\par} &
    \includegraphics[width=\linewidth]{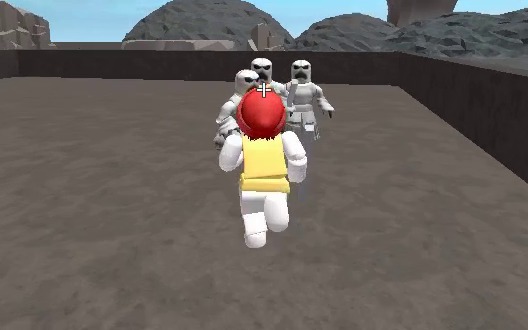}\newline{\scriptsize\textbf{Volcanic}, Skyler, FOV 67$^\circ$\par}{\tiny\itshape ``Fight off the group of sword fighters.''\par} \\[3pt]
\end{tabular}
\end{minipage}}
  \caption{\red{\textbf{Diversity within a generated skill.} Eight verified Melee arena episodes share a combat mechanic while varying the world theme, characters, camera, dynamics, and instruction wording. Player and enemy appearances are sampled from 20 and 9 \anon{}{Roblox }avatar bundles, respectively; camera FOV ranges from 62--78$^\circ$. Themes vary terrain, props, sky, and lighting. Arena size, enemy speeds, and control gains also vary across episodes. Labels give the theme, player avatar, FOV, and instruction.}}
  \label{fig:diversity}
\end{figure*}

\let\supplementaryparagraph\paragraph
\makeatletter
\renewcommand\paragraph{\@startsection{paragraph}{4}{\z@}{3pt plus 1pt minus 1pt}{-1em}{\normalfont\normalsize\bfseries}}
\makeatother
\section{Related Work}
\label{sec:related}

\subsection{Visual Game-Playing Agents}

\red{Visual game-playing agents map rendered frames to actions through keyboard, mouse, or gamepad interfaces.}
\red{In Minecraft, early policies learn behavioral priors from video~\cite{baker2022vpt,lifshitz2023steve1,cai2025rocket1}. More recent methods post-train VLMs to produce keyboard-and-mouse actions~\cite{wang2024omnijarvis,li2025jarvisvla} or combine high-level planners with frame-level controllers~\cite{wang2025openha,tang2026histeve,wang2026taskframe}.}
\red{Beyond Minecraft, generalist agents learn across virtual worlds through unified control interfaces~\cite{sima2024scaling,simateam2025sima2,tan2024cradle,qin2025uitars,wang2025gametars,tan2025lumine}, large-scale behavior cloning~\cite{yue2025pixels2play,magne2026nitrogen,yue2026openp2p}, joint world-action modeling~\cite{guo2026gamewam}, and continued pretraining across worlds~\cite{zhang2026scaling}.}
\red{We focus on the source of post-training supervision: programmatic experts in automatically generated games provide demonstrations without newly collected human gameplay or target-game RL.}

\subsection{\red{Learning in Generated Environments}}

\red{Procedural generation~\cite{cobbe2020procgen,chevalier2019babyai} and generative world models~\cite{bruce2024genie} diversify training environments. LLM-driven pipelines further generate executable tasks and environments for embodied agents~\cite{zala2024envgen,wang2024gensim,wang2024robogen}, tool use~\cite{wang2026awm}, and game engines~\cite{chi2026gamedevbench,luo2026gamecraft,zhou2026unitybench}.}
\red{Gym-Anything-RT~\cite{aggarwal2026gymanythingrt} uses coding agents to build real-time minigames for evaluation. Gym-Anything for Robotics~\cite{aggarwal2026gymanythingrobotics} audits generated robotics simulations and trains VLAs with privileged code policies.}
\red{Our pipeline generates games for specified gameplay skills, audits them to reject trivial or instruction-agnostic tasks, and records scripted teachers acting in real time at the student's control rate.}
\red{The resulting demonstrations train visual policies by behavior cloning, without VLM annotation.}

\subsection{Game Datasets and Benchmarks}

\red{Game-agent training data includes Minecraft recordings and knowledge corpora~\cite{guss2019minerl,baker2022vpt,fan2022minedojo}; multi-game and desktop trajectories from Pixels2Play (Open-P2P)~\cite{yue2025pixels2play,yue2026openp2p}, D2E~\cite{choi2026d2e}, Gaming-500-Hours~\cite{markov2026gaming500}, and NitroGen~\cite{magne2026nitrogen}; and synthetic captures from platforms such as Isekai3D SDG~\cite{karp2026isekai3d}.}
\red{Public benchmarks cover first-person combat in ViZDoom~\cite{kempka2016vizdoom}; Minecraft tasks in MineStudio~\cite{cai2024minestudio}, MCU~\cite{zheng2025mcu}, and OpenHA~\cite{wang2025openha}; classic console and DOS games in VideoGameBench~\cite{zhang2025videogamebench}; verifiable browser games in GameWorld~\cite{ouyang2026gameworld}; and real-time UE5 games in OmniGameArena~\cite{lin2026omnigamearena}.}
\red{Our pipeline generates aligned instructions, frames, and actions with programmatic verification. We evaluate the resulting policies on public benchmarks to measure instruction following and cross-engine transfer.}

\section{Method}
\label{sec:method}

\red{\Cref{fig:pipeline} summarizes the framework. A coding agent turns a gameplay skill specification into a game with varied scenes, instructions, and task verifiers. Successful scripted-teacher rollouts provide demonstrations for behavior cloning. We describe policy training in \cref{sec:method:problem} and game generation in \cref{sec:method:generation}.}


\subsection{Instruction-Following Training}
\label{sec:method:problem}
\label{sec:method:training}

\paragraph{Student and teacher.}
\red{Let $t$ index control decisions at 200\,ms intervals. We denote the engine state by $s_t$, the rendered observation by $o_t = R(s_t)$, the instruction by $g$, and a keyboard-and-mouse action chunk by $a_t = (k_t, m_t)$.}
\red{The student policy, \textsc{Script2Play}, is $\pi_\theta(a_t \mid g, o_t)$, conditioned only on the instruction and current frame, with no access to engine state.}
\red{The coding agent writes a teacher $\pi^*(a_t \mid s_t, g)$ for each game.}
It reads character poses and velocities, target coordinates, enemy and projectile states, and collision geometry.
\red{The teacher selects actions through pathfinding, predictive aiming, finite-state control, or timing rules, depending on the task.}
\red{It uses the same keyboard-and-mouse interface and control rate as the student.}

\paragraph{Teacher rollouts.}
A task $\tau = (s_0, g, V)$ consists of an initial scene $s_0$, an instruction $g$, and a verifier $V$.
\red{The verifier checks task completion within a time budget $T_\tau$ and detects prohibited events, such as reaching the wrong target or falling, from engine state.}
A teacher rollout with $L_\tau$ control decisions is
\begin{equation}
  \xi_\tau = \big((g, o_t, a^*_t)\big)_{t=1}^{L_\tau},
  \label{eq:trajectory}
\end{equation}
and the training corpus retains only successful rollouts: $\mathcal{D} = \{\xi_\tau : V(\xi_\tau)=1\}$.
\red{We record observations at 20\,fps and align them with 200\,ms action windows.}
\red{The task specification provides instructions, the teacher provides actions, and the verifier provides success labels. This process requires neither human gameplay demonstrations nor retrospective VLM annotation.}

\paragraph{Behavior cloning.}
\red{We initialize the student from a pretrained open-weight VLM and train it to predict teacher actions. The unweighted gameplay loss is}
\begin{equation}
  \red{\mathcal{L}_{\mathrm{BC}}(\theta) = -\mathbb{E}_{\tau,\, t} \big[ \log \pi_\theta(a^*_t \mid g, o_t) \big],}
  \label{eq:bc}
\end{equation}
where $\tau$ indexes a retained rollout and $t$ a control decision within it.
\red{Our \textsc{Script2Play} policies use behavior cloning on synthetic gameplay.}
\red{To emphasize control changes, we upweight the token loss for windows whose held-key set differs from the preceding window, by a factor of 5.}
\red{Generic visual question answering, optical character recognition, and text examples occupy 15\% of each batch to help retain general capabilities. These examples contain no gameplay.}
\red{At inference, the policy acts in closed loop in each benchmark's native engine, mapping instructions and frames to keyboard and mouse actions.}
\red{We adapt Lumine's~\cite{tan2025lumine} text-based action format, autoregressively generating:} $\Delta X\ \Delta Y\,;\, K_1\,;\, K_2\,;\, K_3\,;\, K_4$.
The mouse displacement $(\Delta X,\Delta Y)$ is distributed across the 200\,ms window, while $K_1$ through $K_4$ specify the keys and mouse buttons held in its four successive 50\,ms intervals. Unlisted keys are released.

\subsection{Data Generation}
\label{sec:method:generation}
\label{sec:method:teacher}

\paragraph{From specification to game.}
\red{\Cref{fig:generation} illustrates the generation and verification process for one game.}
A specification has five fields: the gameplay skill, camera, dynamics ranges, controls, and task contract (\red{Supplementary \cref*{tab:spec}}).
\red{The task contract specifies the instructed target and the events that invalidate an episode.}
The coding agent implements the scene, interactive mechanics, task verifier, and teacher using a shared runtime library (\red{Supplementary \cref*{sec:supp:agent}}).
\red{The agent also writes a scene generator that varies layouts, spawn points, targets, dynamics, and assets.}
\red{To check that success depends on the instruction, we create task pairs with identical scenes but different target choices.}
\red{For example, in \cref{fig:generation}, shooting all blue targets must fail when the instruction asks for all red targets, and vice versa.}
\red{Randomization varies the world theme, player and NPC appearances, camera, layout, dynamics, and instruction wording while preserving the gameplay mechanic (\cref{fig:diversity}).}
\red{This variation aims to reduce dependence on a particular visual setting.}

\paragraph{Audit and validation.}
\red{We use a Python implementation of the game rules for preliminary teacher and verifier tests, followed by validation in the game engine.}
\red{Before its rollouts enter $\mathcal{D}$, each game must pass five automated audit gates (details in Supplementary \cref*{tab:gates}):}
\begin{itemize}
  \setlength{\itemsep}{0pt}\setlength{\parskip}{0pt}
  \item \textbf{G1 Static:} the code compiles, and paired tasks share the same scene.
  \item \textbf{G2 Solvable:} \red{the teacher succeeds in at least 95\% of Python test episodes, both with and without added latency.}
  \item \textbf{G3 Non-trivial:} idle, hold-forward, and random-key policies rarely succeed, and a teacher pursuing the other task's target also rarely succeeds on the instructed task.
  \item \textbf{G4 Engine parity:} \red{we rerun the same teacher in the game engine and require a success rate within five percentage points of the Python tests (G2).}
  \item \textbf{G5 Visual:} recorded frames are non-blank and change over the episode.
\end{itemize}
G3 screens for two failure modes: tasks that can be solved without observing the scene, and verifiers that accept the wrong instructed target.
\red{For recoverable failures, the agent adjusts task tolerances to improve teacher success or remove trivial solutions, then repeats the audit for up to three repair rounds.}
Games that fail G1 or the wrong-target check are discarded (\red{Supplementary \cref*{alg:pipeline}}).

\begin{figure*}[t]
  \centering
  \setlength{\tabcolsep}{1.5pt}
  \renewcommand{\arraystretch}{0.6}
  \scriptsize
\begin{tabular}{@{}>{\raggedright\arraybackslash}p{0.235\linewidth}*{5}{p{0.147\linewidth}}@{}}
\vspace{0pt}{\textbf{OpenHA: combat}\par ``Hunt the pillager.''\par} &
\vspace{0pt}{\centering\includegraphics[width=0.88\linewidth]{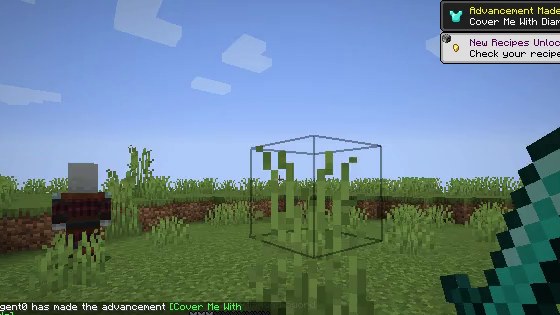}\par step 1~\texttt{(-150,11)}\par\texttt{W\,|\,W\,|\,W\,|\,W}\par} &
\vspace{0pt}{\centering\includegraphics[width=0.88\linewidth]{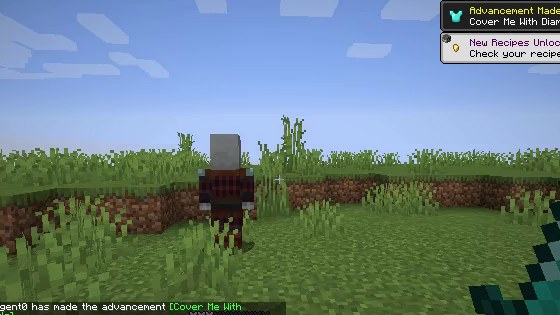}\par step 4~\texttt{(-113,1)}\par\texttt{W\,|\,W\,|\,W\,|\,W}\par} &
\vspace{0pt}{\centering\includegraphics[width=0.88\linewidth]{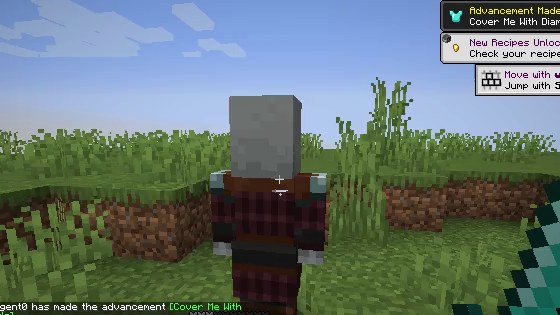}\par step 7~\texttt{(-23,34)}\par\texttt{W+LB\,|\,W\,|\,W\,|\,W}\par} &
\vspace{0pt}{\centering\includegraphics[width=0.88\linewidth]{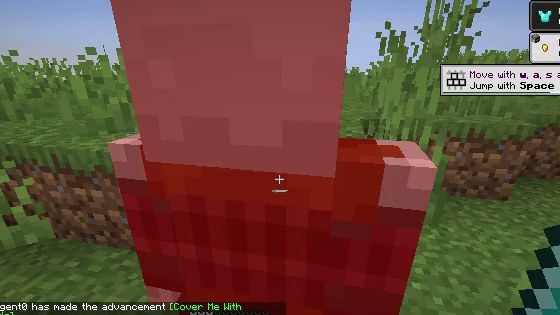}\par step 10~\texttt{(-1,-7)}\par\texttt{LB\,|\,$\cdot$\,|\,$\cdot$\,|\,$\cdot$}\par} &
\vspace{0pt}{\centering\includegraphics[width=0.88\linewidth]{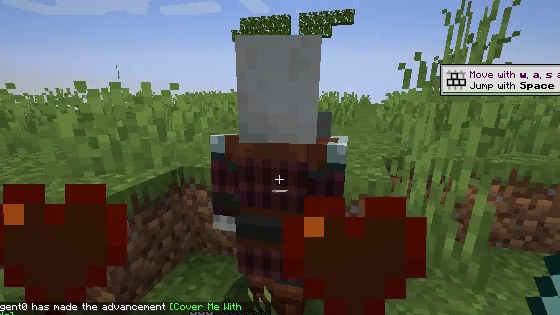}\par step 13~\texttt{(0,5)}\par\texttt{LB\,|\,$\cdot$\,|\,$\cdot$\,|\,$\cdot$}\par} \\[2pt]
\vspace{0pt}{\textbf{OpenHA: embodied}\par ``Mine the mushroom stem in the forest.''\par} &
\vspace{0pt}{\centering\includegraphics[width=0.88\linewidth]{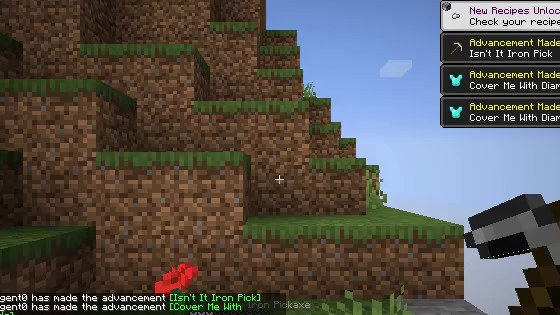}\par step 0~\texttt{(-106,10)}\par\texttt{W\,|\,W\,|\,W\,|\,W}\par} &
\vspace{0pt}{\centering\includegraphics[width=0.88\linewidth]{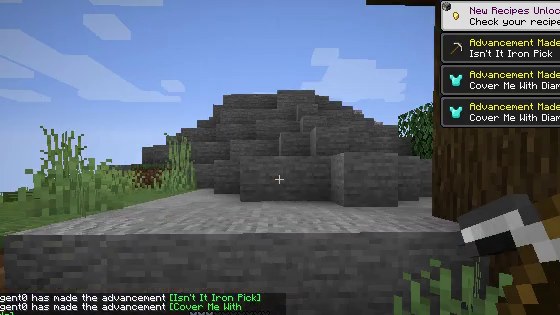}\par step 4~\texttt{(-206,167)}\par\texttt{$\cdot$\,|\,$\cdot$\,|\,$\cdot$\,|\,$\cdot$}\par} &
\vspace{0pt}{\centering\includegraphics[width=0.88\linewidth]{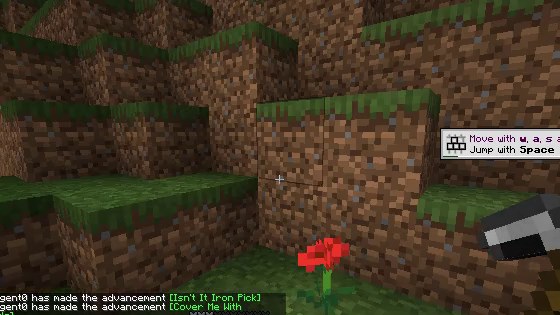}\par step 8~\texttt{(-219,0)}\par\texttt{$\cdot$\,|\,$\cdot$\,|\,$\cdot$\,|\,$\cdot$}\par} &
\vspace{0pt}{\centering\includegraphics[width=0.88\linewidth]{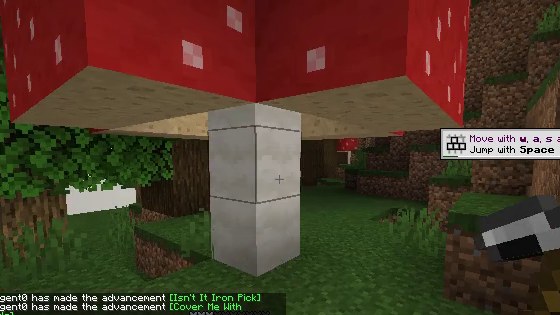}\par step 11~\texttt{(-18,3)}\par\texttt{LB\,|\,LB\,|\,LB\,|\,LB}\par} &
\vspace{0pt}{\centering\includegraphics[width=0.88\linewidth]{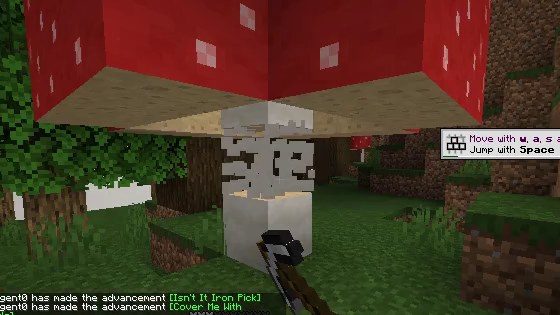}\par step 13~\texttt{(0,0)}\par\texttt{LB\,|\,LB\,|\,LB\,|\,LB}\par} \\[2pt]
\vspace{0pt}{\textbf{OpenHA: GUI}\par ``Craft golden\_leggings.''\par} &
\vspace{0pt}{\centering\includegraphics[width=0.88\linewidth]{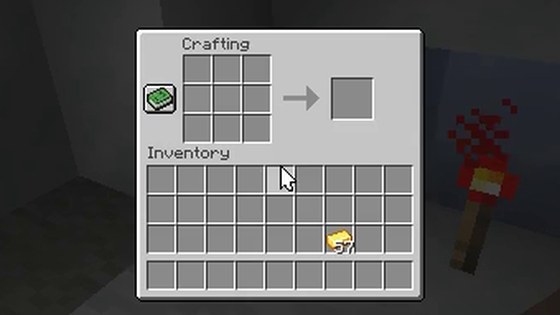}\par step 0~\texttt{(37,49)}\par\texttt{$\cdot$\,|\,$\cdot$\,|\,$\cdot$\,|\,RB}\par} &
\vspace{0pt}{\centering\includegraphics[width=0.88\linewidth]{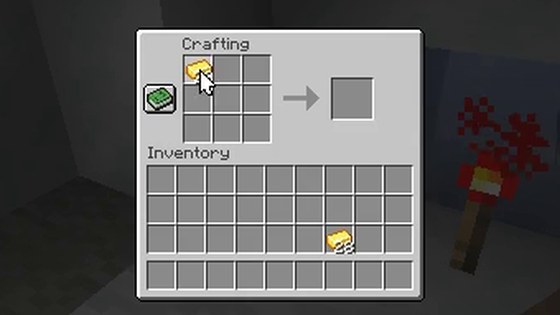}\par step 2~\texttt{(20,-3)}\par\texttt{$\cdot$\,|\,$\cdot$\,|\,RB\,|\,$\cdot$}\par} &
\vspace{0pt}{\centering\includegraphics[width=0.88\linewidth]{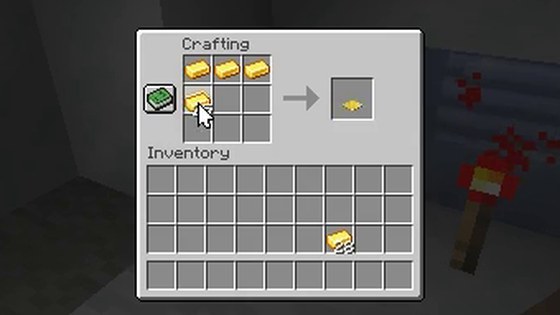}\par step 5~\texttt{(36,-4)}\par\texttt{$\cdot$\,|\,$\cdot$\,|\,RB\,|\,$\cdot$}\par} &
\vspace{0pt}{\centering\includegraphics[width=0.88\linewidth]{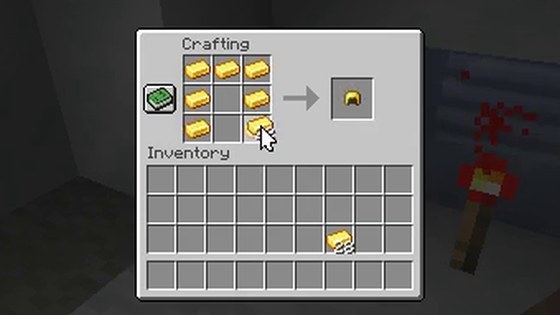}\par step 8~\texttt{(49,72)}\par\texttt{$\cdot$\,|\,$\cdot$\,|\,LB\,|\,$\cdot$}\par} &
\vspace{0pt}{\centering\includegraphics[width=0.88\linewidth]{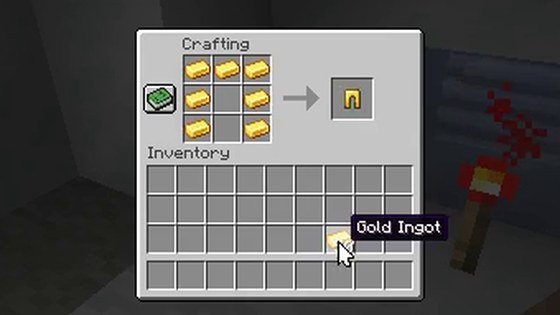}\par step 10~\texttt{(6,-88)}\par\texttt{$\cdot$\,|\,$\cdot$\,|\,$\cdot$\,|\,Shift+LB}\par} \\[2pt]
\vspace{0pt}{\textbf{OmniGameArena: ObstacleRun3D}\par Run across the floating platforms and jump the gaps to the finish.\par} &
\vspace{0pt}{\centering\includegraphics[width=0.88\linewidth]{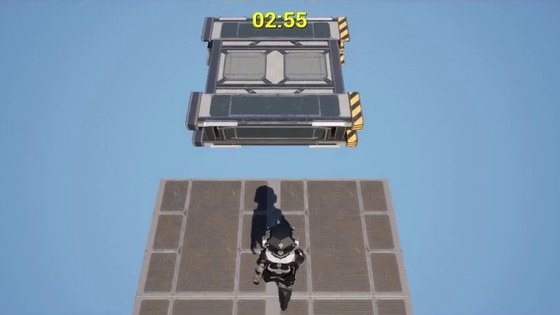}\par step 10~\texttt{(0,0)}\par\texttt{W\,|\,W\,|\,W\,|\,W}\par} &
\vspace{0pt}{\centering\includegraphics[width=0.88\linewidth]{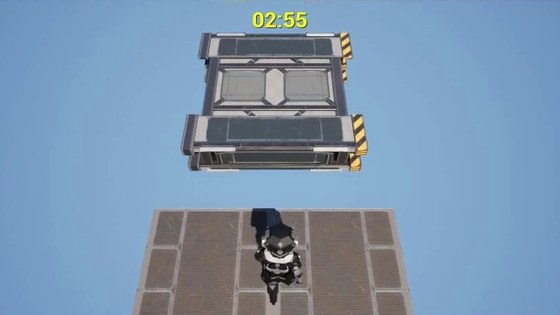}\par step 12~\texttt{(0,0)}\par\texttt{W\,|\,W\,|\,W+Spc\,|\,W+Spc}\par} &
\vspace{0pt}{\centering\includegraphics[width=0.88\linewidth]{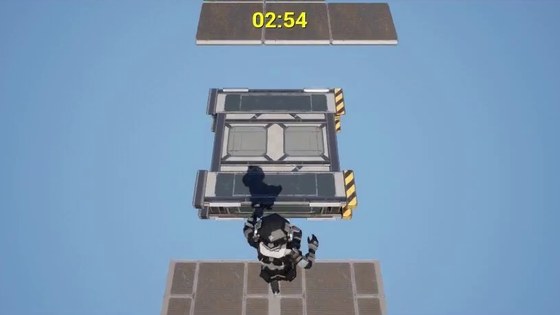}\par step 14~\texttt{(0,0)}\par\texttt{W\,|\,W+Spc\,|\,W+Spc\,|\,W}\par} &
\vspace{0pt}{\centering\includegraphics[width=0.88\linewidth]{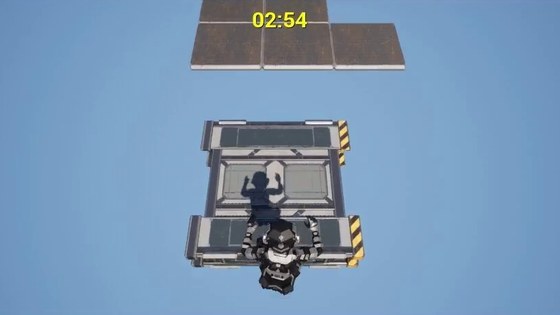}\par step 16~\texttt{(0,0)}\par\texttt{W\,|\,W\,|\,W\,|\,W}\par} &
\vspace{0pt}{\centering\includegraphics[width=0.88\linewidth]{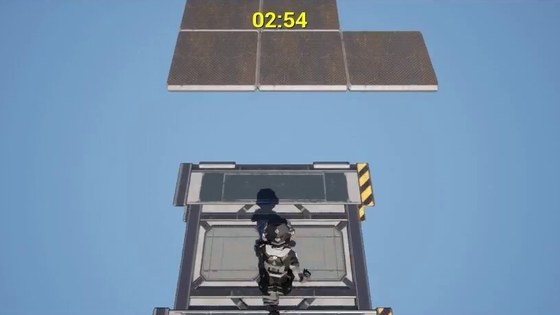}\par step 18~\texttt{(0,0)}\par\texttt{W\,|\,W\,|\,W\,|\,W}\par} \\[2pt]
\vspace{0pt}{\textbf{OmniGameArena: CueChase}\par ``Pass your Feather to Sarah, and she will reward you with a Mushroom.'' (on screen)\par} &
\vspace{0pt}{\centering\includegraphics[width=0.88\linewidth]{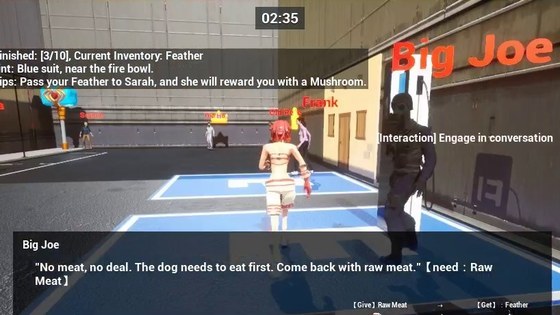}\par step 161~\texttt{(1,0)}\par\texttt{W\,|\,W\,|\,W\,|\,W}\par} &
\vspace{0pt}{\centering\includegraphics[width=0.88\linewidth]{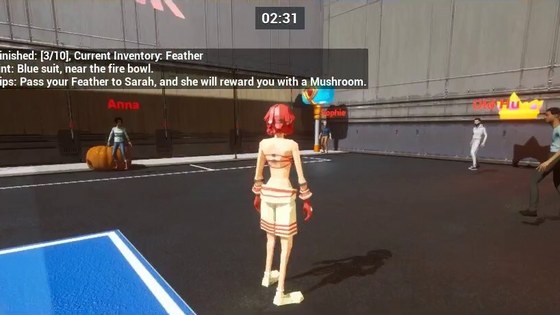}\par step 184~\texttt{(-153,0)}\par\texttt{$\cdot$\,|\,$\cdot$\,|\,$\cdot$\,|\,$\cdot$}\par} &
\vspace{0pt}{\centering\includegraphics[width=0.88\linewidth]{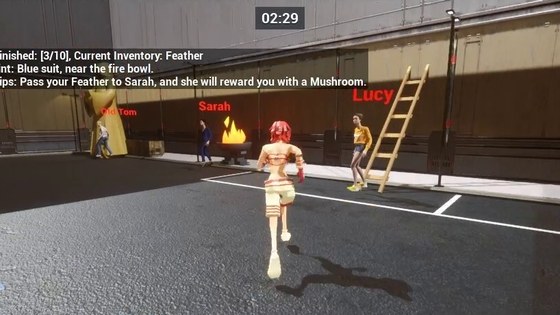}\par step 196~\texttt{(0,0)}\par\texttt{W\,|\,W\,|\,W\,|\,W}\par} &
\vspace{0pt}{\centering\includegraphics[width=0.88\linewidth]{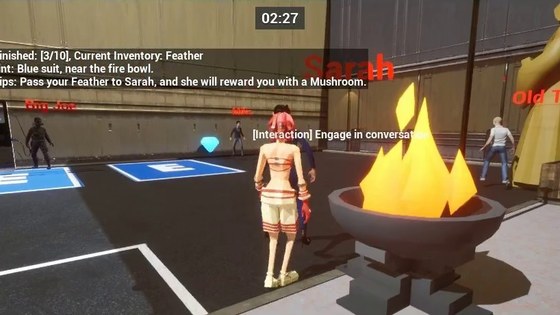}\par step 209~\texttt{(0,0)}\par\texttt{W\,|\,W\,|\,W\,|\,W}\par} &
\vspace{0pt}{\centering\includegraphics[width=0.88\linewidth]{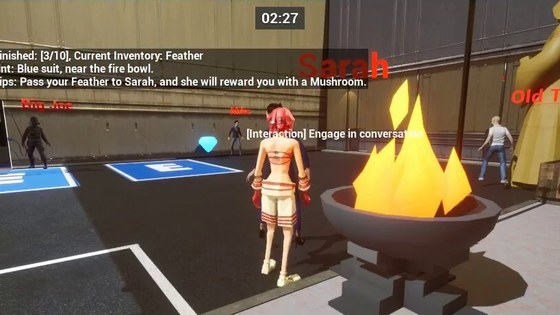}\par step 211~\texttt{(0,0)}\par\texttt{F\,|\,F\,|\,$\cdot$\,|\,$\cdot$}\par} \\
\end{tabular}

  \caption{\red{\textbf{Closed-loop policy actions across engines.} Each row shows five sampled observation--action pairs from a successful \textsc{Script2Play} episode. Actions contain mouse displacement $(dx,dy)$ in pixels and four 50\,ms key groups ($\cdot$: no key; LB/RB: left/right mouse button; Spc: space). Each step covers 200\,ms.}}
  \label{fig:io}
\end{figure*}

\begin{table*}[t]
  \centering
  \caption{\textbf{Instruction following on OpenHA} (success rate, \%, with 95\% confidence intervals in brackets for \textsc{Script2Play}; higher is better). Total data denotes the listed gameplay training scale; MC data denotes its Minecraft component. STEVE-1 includes inherited VPT pretraining. \textsc{Script2Play} uses 150.3\,h in total: \red{26.9\,h from Minecraft and 123.4\,h from \anon{an in-house game engine}{Roblox}}. Bold marks the best score separately within commercial VLMs and trained policies (including ours). $^\dagger$Approximate hour equivalents inferred from reported frame or token counts; see \red{Supplementary \cref*{sec:supp:exp:accounting}}.}
  \label{tab:openha}
  \scriptsize
  \setlength{\tabcolsep}{3pt}
  \renewcommand{\arraystretch}{0.95}
  \newcommand{\oci}[2]{\makecell[c]{#1\\[-2.5pt]{\fontsize{6}{6.5}\selectfont[#2]}}}
  \begin{tabular*}{\textwidth}{@{\extracolsep{\fill}}llllcccc@{}}
    \toprule
    Method & Training data & Total data & MC data & Embodied & GUI & Combat & Avg \\
    \midrule
    \multicolumn{8}{@{}l}{\textit{Commercial VLMs (low reasoning effort)}} \\
    \quad Claude Opus 5.5 & -- & -- & -- & 72.9 & 80.6 & 53.1 & 68.9 \\
    \quad GPT-6.1-Sol & -- & -- & -- & 68.7 & 84.9 & 69.4 & 74.3 \\
    \quad Gemini 4 Argon & -- & -- & -- & \textbf{80.7} & \textbf{96.0} & \textbf{71.4} & \textbf{82.7} \\
    \midrule
    \multicolumn{8}{@{}l}{\textit{Trained Game Policies}} \\
    \quad VPT~\cite{baker2022vpt} & Human play & $\sim$70{,}000\,h & $\sim$70{,}000\,h & 6.0 & 0.8 & 3.6 & 3.5 \\
    \quad STEVE-1~\cite{lifshitz2023steve1} & Human + VPT rollouts & $\sim$70{,}750\,h$^\dagger$ & $\sim$70{,}750\,h$^\dagger$ & 8.0 & 3.2 & 3.9 & 5.0 \\
    \quad ROCKET-1~\cite{cai2025rocket1} & Human play & $\sim$22{,}000\,h$^\dagger$ & $\sim$22{,}000\,h$^\dagger$ & 18.9 & 0.0 & 27.9 & 15.6 \\
    \quad JARVIS-VLA~\cite{li2025jarvisvla} & Human play & $\sim$192\,h$^\dagger$ & $\sim$192\,h$^\dagger$ & 30.0 & 25.1 & 18.5 & 24.5 \\
    \quad OpenHA~\cite{wang2025openha} & Human play & $\sim$145\,h$^\dagger$ & $\sim$145\,h$^\dagger$ & 30.1 & 32.5 & 31.9 & 31.5 \\
    \quad UI-TARS-1.5~\cite{qin2025uitars} & Human play + RL gym & $\sim$2{,}600\,h$^\dagger$ & $\sim$600\,h$^\dagger$ & 42.1 & 36.7 & 31.0 & 36.6 \\
    \quad HiSTEVE~\cite{tang2026histeve} & Human play + RL gym & $\sim$200\,h$^\dagger$ + RL & $\sim$200\,h$^\dagger$ + RL & 46.0 & 27.2 & 39.0 & 37.4 \\
    \quad Game-TARS Dense 7B~\cite{wang2025gametars} & Human play + RL gym & $\sim$20{,}000\,h & $\sim$560\,h$^\dagger$ & 50.4 & 39.1 & 38.1 & 42.5 \\
    \quad GameWAM~\cite{guo2026gamewam} & Human play + scripted GUI & $\sim$300\,h$^\dagger$ & $\sim$300\,h$^\dagger$ & 47.5 & 60.0 & 32.2 & 46.6 \\
    \quad Game-TARS MoE-mini~\cite{wang2025gametars} & Human play + RL gym & $\sim$22{,}000\,h & $\sim$560\,h$^\dagger$ & \textbf{72.0} & 55.4 & 66.1 & 64.5 \\
    \midrule
    \multicolumn{8}{@{}l}{\textit{Our method}} \\
    \quad \textsc{Script2Play} (4B) & Scripted teacher & 150.3\,h & 26.9\,h & \oci{63.9}{58.7, 69.1} & \oci{\textbf{66.4}}{63.7, 69.1} & \oci{\textbf{78.6}}{70.3, 86.9} & \oci{\textbf{69.6}}{66.3, 73.0} \\
    \bottomrule
  \end{tabular*}

\end{table*}

\begin{table*}[t]
  \centering
  \caption{\textbf{Transfer to five solo games in OmniGameArena.} Scores use the benchmark's per-game normalized metrics (with 95\% confidence intervals in brackets for locally evaluated policies) and their mean (higher is better), under the time-paused protocol~\cite{lin2026omnigamearena,zhang2026scaling}. \textsc{Script2Play} receives no UE5 gameplay during post-training. Bold marks the best score in each column. Training scales describe the respective gameplay corpora, not matched compute budgets.}
  \label{tab:omni}
  \scriptsize
  \setlength{\tabcolsep}{0pt}
  \renewcommand{\arraystretch}{0.95}
  \newcommand{\oci}[2]{\makecell[c]{#1\\[-2.5pt]{\fontsize{6}{6.5}\selectfont[#2]}}}
  \begin{tabular*}{\textwidth}{@{\extracolsep{\fill}}lllcccccc@{}}
    \toprule
    Agent & Training data & Data scale & \makecell{Obstacle\\Run3D} & \makecell{Last\\Stand} & \makecell{Monster\\Shoot} & \makecell{Scene\\Escape} & \makecell{Cue\\Chase} & Mean \\
    \midrule
    \multicolumn{9}{@{}l}{\textit{Commercial VLMs}} \\
    \quad Claude Opus 4.7~\cite{lin2026omnigamearena} & -- & -- & 0.094 & 0.308 & 0.400 & 0.460 & 0.380 & 0.328 \\
    \quad Claude Opus 4.6~\cite{lin2026omnigamearena} & -- & -- & 0.172 & 0.147 & 0.362 & 0.540 & 0.840 & 0.412 \\
    \quad Claude Sonnet 4.6~\cite{lin2026omnigamearena} & -- & -- & 0.200 & 0.144 & 0.166 & 0.440 & 0.440 & 0.278 \\
    \quad GPT-5.5~\cite{lin2026omnigamearena} & -- & -- & 0.133 & 0.416 & 0.464 & \textbf{0.720} & 0.580 & 0.463 \\
    \quad GPT-5.4~\cite{lin2026omnigamearena} & -- & -- & 0.089 & 0.148 & 0.326 & 0.680 & 0.300 & 0.309 \\
    \quad Gemini 3.1 Pro Preview~\cite{lin2026omnigamearena} & -- & -- & 0.165 & 0.230 & 0.710 & 0.660 & 0.600 & 0.473 \\
    \quad Gemini 3.1 Flash-Lite Preview~\cite{lin2026omnigamearena} & -- & -- & 0.097 & 0.122 & 0.182 & 0.300 & 0.440 & 0.228 \\
    \quad Kimi K2.5~\cite{lin2026omnigamearena} & -- & -- & 0.075 & 0.232 & 0.290 & 0.220 & 0.140 & 0.191 \\
    \midrule
    \multicolumn{9}{@{}l}{\textit{Open-weight VLMs}} \\
    \quad Qwen3.5-397B-A17B~\cite{lin2026omnigamearena} & -- & -- & 0.112 & 0.106 & 0.072 & 0.200 & 0.040 & 0.106 \\
    \quad Qwen3.5-122B-A10B~\cite{lin2026omnigamearena} & -- & -- & 0.034 & 0.093 & 0.024 & 0.060 & 0.040 & 0.050 \\
    \midrule
    \multicolumn{9}{@{}l}{\textit{Trained Game Policies}} \\
    \quad NitroGen (gamepad)~\cite{magne2026nitrogen} & Human play & 40{,}000\,h & 0.065 & 0.100 & 0.014 & 0.120 & 0.020 & 0.064 \\
    \quad Open-P2P (kbd-mouse)~\cite{yue2026openp2p} & Human play & 8{,}300+\,h & 0.023 & 0.063 & 0.000 & 0.100 & 0.000 & 0.037 \\
    \quad GameScaling-0.8B-S3~\cite{zhang2026scaling} & Human play & 1{,}600\,h & \oci{0.182}{0.000, 0.416} & \oci{0.102}{0.068, 0.137} & \oci{0.316}{0.073, 0.559} & \oci{0.200}{0.048, 0.352} & \oci{0.040}{0.000, 0.108} & 0.168 \\
    \quad GameScaling-2B-S3~\cite{zhang2026scaling} & Human play & 1{,}600\,h & \oci{0.097}{0.000, 0.236} & \oci{0.102}{0.060, 0.144} & \oci{0.314}{0.000, 0.724} & \oci{0.160}{0.000, 0.386} & \oci{0.000}{0.000, 0.000} & 0.135 \\
    \quad GameScaling-9B-S3~\cite{zhang2026scaling} & Human play & 1{,}600\,h & \oci{0.083}{0.038, 0.129} & \oci{0.080}{0.053, 0.107} & \oci{0.780}{0.304, 1.000} & \oci{0.520}{0.298, 0.742} & \oci{0.340}{0.152, 0.528} & 0.361 \\
    \midrule
    \multicolumn{9}{@{}l}{\textit{Our method}} \\
    \quad \textsc{Script2Play} (4B) & Scripted teacher & 150.3\,h & \oci{\textbf{0.374}}{0.104, 0.644} & \oci{\textbf{0.436}}{0.387, 0.485} & \oci{\textbf{0.972}}{0.924, 1.000} & \oci{0.420}{0.198, 0.642} & \textbf{1.000} & \oci{\textbf{0.640}}{0.582, 0.699} \\
    \bottomrule
  \end{tabular*}

\end{table*}

\begin{table*}[t]
  \centering
  \begin{minipage}[t]{0.49\textwidth}
  \vspace{0pt}
  \centering
  \caption{\textbf{Transfer to ViZDoom} (mean reward over 50 episodes per map; higher is better). GameWAM uses target-domain training; the other policies transfer zero-shot. $^\dagger$Approximate data scale; see \red{Supplementary \cref*{sec:supp:exp:accounting}} for accounting \red{details}.}
  \label{tab:vizdoom}
  \fontsize{7.5}{8.5}\selectfont
  \setlength{\tabcolsep}{0.7pt}
  \begin{tabular}{@{}llccccc@{}}
    \toprule
    Method & Data scale & \makecell{Battle\\1} & \makecell{Battle\\2} & \makecell{Defend\\Line} & \makecell{Defend\\Center} & Mean \\
    \midrule
    \multicolumn{7}{@{}l}{\textit{In-domain (trained on ViZDoom APPO)}} \\
    \quad GameWAM~\cite{guo2026gamewam} & $\sim$300\,h$^\dagger$ & 43.12 & 19.02 & 23.52 & 15.36 & 25.26 \\
    \midrule
    \multicolumn{7}{@{}l}{\textit{Zero-shot (no ViZDoom training)}} \\
    \quad \makecell[l]{Game-TARS\\MoE-mini}~\cite{wang2025gametars} & $\sim$22{,}000\,h & 18.87 & 10.04 & 15.24 & 14.00 & 14.54 \\
    \quad o4-mini~\cite{guo2026gamewam} & -- & 7.85 & 6.74 & 10.24 & 11.39 & 9.06 \\
    \quad GPT-5~\cite{guo2026gamewam} & -- & 12.59 & 8.99 & 10.24 & 13.46 & 11.32 \\
    \quad Gemini 2.5 Pro~\cite{guo2026gamewam} & -- & 7.51 & 10.04 & 15.24 & 5.27 & 9.52 \\
    \quad Claude 4 Sonnet~\cite{guo2026gamewam} & -- & 7.34 & 7.52 & 10.24 & 5.56 & 7.67 \\
    \midrule
    \multicolumn{7}{@{}l}{\textit{Our method (scripted teacher)}} \\
    \quad \textsc{Script2Play} (4B) & 150.3\,h & 17.00 & 4.08 & 22.94 & 7.66 & 12.92 \\
    \bottomrule
  \end{tabular}

  \end{minipage}\hfill
  \begin{minipage}[t]{0.49\textwidth}
  \vspace{0pt}
  \centering
  \caption{\textbf{Supervision comparison with Qwen2.5-VL-7B.} Our 7B variant uses behavior cloning on 26.9\,h of Minecraft data, independently of the main 4B model. Brackets show 95\% confidence intervals; bold marks column bests. HiSTEVE weights are unavailable for (b). $^\dagger$Approximate hour equivalents (\red{Supplementary \cref*{sec:supp:exp:accounting}}).}
  \label{tab:abl:qwen25}
  \fontsize{7.5}{8.5}\selectfont
  \renewcommand{\arraystretch}{1.0}
  \newcommand{\oci}[2]{\makecell[c]{#1\\[-3pt]{\tiny[#2]}}}
  \centering
  \setlength{\tabcolsep}{0pt}
  \begin{tabular*}{\linewidth}{@{\extracolsep{\fill}}lllcccc@{}}
    \toprule
    \multicolumn{7}{@{}l}{\textit{(a) In-domain: OpenHA (Minecraft, success rate \%)}} \\
    \midrule
    Supervision & Method & MC data & Emb. & GUI & Combat & Avg \\
    \midrule
    Human play & OpenHA~\cite{wang2025openha} & $\sim$145\,h$^\dagger$ & 30.1 & 32.5 & 31.9 & 31.5 \\
    Human + RL & HiSTEVE~\cite{tang2026histeve} & $\sim$200\,h$^\dagger$ + RL & 46.0 & 27.2 & 39.0 & 37.4 \\
    Scripted & \makecell[l]{\textsc{Script2Play}\\(7B, MC only)} & 26.9\,h & \oci{\textbf{49.7}}{44.3, 55.1} & \oci{\textbf{39.0}}{36.2, 41.8} & \oci{\textbf{49.0}}{38.9, 59.1} & \textbf{45.9} \\
    \bottomrule
  \end{tabular*}
  \par\smallskip
  \centering
  \setlength{\tabcolsep}{1pt}
  \begin{tabular*}{\linewidth}{@{\extracolsep{\fill}}lcccccc@{}}
    \toprule
    \multicolumn{7}{@{}l}{\textit{(b) Out-of-distribution: OmniGameArena (UE5, normalized score)}} \\
    \midrule
    Method & \makecell{Obstacle\\Run3D} & \makecell{Last\\Stand} & \makecell{Monster\\Shoot} & \makecell{Scene\\Escape} & \makecell{Cue\\Chase} & Mean \\
    \midrule
    OpenHA~\cite{wang2025openha} & 0.021 & \textbf{0.082} & \textbf{0.010} & 0.000 & 0.000 & 0.023 \\
    \makecell[l]{\textsc{Script2Play}\\(7B, MC only)} & \oci{\textbf{0.033}}{0.000, 0.073} & \oci{0.066}{0.041, 0.090} & \oci{0.000}{0.000, 0.000} & \oci{\textbf{0.040}}{0.000, 0.108} & \oci{0.000}{0.000, 0.000} & \textbf{0.028} \\
    \bottomrule
  \end{tabular*}

  \end{minipage}
\end{table*}

\begin{table}[t]
  \revisionfloat
  \centering
  \caption{\textbf{Skill-category ablation on OmniGameArena.} Normalized scores over 10 episodes (mean with 95\% confidence intervals; higher is better). C: Combat \& Shooting; T: Traversal \& Parkour; D: Dodging \& Survival. Bold marks the best ablation arm per column. Reference \textsc{Script2Play} uses 8{,}000 training steps on the full mixture and 5 evaluation episodes per game.}
  \label{tab:skills}
  \fontsize{8}{9}\selectfont
  \setlength{\tabcolsep}{3pt}
  \renewcommand{\arraystretch}{1.10}
  \newcommand{\oci}[2]{\makecell[c]{#1\\[-2.5pt]{\scriptsize[#2]}}}
  \begin{tabular}{@{}lcccc@{}}
    \toprule
    Training categories & \makecell{Monster\\Shoot} & \makecell{Obstacle\\Run3D} & \makecell{Last\\Stand} & Mean \\
    \midrule
    C & \oci{0.722}{0.468, 0.976} & \oci{0.066}{0.064, 0.069} & \oci{0.108}{0.079, 0.138} & 0.299 \\
    T & \oci{0.000}{0.000, 0.000} & \oci{0.215}{0.134, 0.295} & \oci{0.139}{0.088, 0.190} & 0.118 \\
    D & \oci{0.000}{0.000, 0.000} & \oci{0.000}{0.000, 0.000} & \oci{0.744}{0.588, 0.900} & 0.248 \\
    C+T & \oci{0.985}{0.956, 1.000} & \oci{\textbf{0.370}}{0.259, 0.481} & \oci{0.118}{0.109, 0.126} & 0.491 \\
    C+D & \oci{\textbf{0.991}}{0.979, 1.000} & \oci{0.014}{0.005, 0.023} & \oci{\textbf{0.812}}{0.663, 0.961} & \textbf{0.606} \\
    T+D & \oci{0.000}{0.000, 0.000} & \oci{0.233}{0.126, 0.341} & \oci{0.427}{0.354, 0.499} & 0.220 \\
    C+T+D & \oci{0.969}{0.948, 0.990} & \oci{0.288}{0.170, 0.407} & \oci{0.428}{0.390, 0.467} & 0.562 \\
    \midrule
    \multicolumn{5}{@{}l}{\textit{Reference}} \\
    \textsc{Script2Play} (4B) & 0.972 & 0.374 & 0.436 & 0.594 \\
    Random~\cite{lin2026omnigamearena} & 0.008 & 0.000 & 0.088 & 0.032 \\
    \bottomrule
  \end{tabular}

\end{table}

\section{Experiments}
\label{sec:exp}
\setcounter{dbltopnumber}{2}

\red{We evaluate instruction following in Minecraft and zero-shot transfer to games in unseen engines. We compare supervision sources with a shared backbone, then compare human and scripted demonstrations with matched data budgets and training settings.}


\subsection{Experimental Setup}
\label{sec:exp:data}

\paragraph{Training data and model.}
\red{We train \textsc{Script2Play} from Qwen3.5-4B~\cite{qwen2026qwen35} on 150.3\,h of verified scripted-teacher gameplay: 26.9\,h from Minecraft and 123.4\,h from \anon{an in-house game engine}{Roblox}.}
\red{The main \textsc{Script2Play} results on all three benchmarks use this single checkpoint (\cref{tab:openha,tab:omni,tab:vizdoom}).}
\red{Training uses 480p frames recorded at 20\,fps and the 200\,ms action representation of \cref{sec:method:training}.}
We freeze the vision encoder and projector and train for \red{8{,}000 steps} on 8 H200 GPUs using bfloat16 and a global batch size of 384 windows.
The learning rate is $2\times10^{-5}$, with 5\% warmup followed by cosine decay to 10\% of its peak value.

\paragraph{Skill coverage.}
\red{We choose the game mixture by surveying games on mainstream platforms and tasks in public benchmarks, then estimating how frequently they require each gameplay skill.}
\red{This skill distribution determines the game-category proportions and demonstration budgets (\cref{fig:data3d:mosaic}). The 123.4\,h from \anon{the in-house game engine}{Roblox} covers 41 generated games in six categories: combat and shooting (56.4\,h), traversal and parkour (21.5\,h), object interaction (20.8\,h), dodging and survival (19.7\,h), GUI operation (3.7\,h), and ball games (1.3\,h).}
\red{The 26.9\,h of Minecraft demonstrations comprise 31{,}494 verified episodes covering mining and workstation use (5.0\,h), crafting and smelting (19.0\,h), and mob combat (2.9\,h).}
Scripted teachers collect these demonstrations in MineStudio~\cite{cai2024minestudio} using 48 custom world seeds disjoint from the evaluation worlds, with randomized spawn locations, tool and inventory loadouts, and surrounding distractors.

\paragraph{Benchmarks and metrics.}
OpenHA~\cite{wang2025openha} evaluates Minecraft instruction following across Embodied, GUI, and Combat tasks; we report success rates for each group and their average. Because Minecraft demonstrations are included in training, OpenHA tests downstream performance within a training engine.
OmniGameArena~\cite{lin2026omnigamearena} and ViZDoom~\cite{kempka2016vizdoom} test cross-engine transfer: neither UE5 nor Doom gameplay is included in our policy post-training.
For OmniGameArena, we evaluate the five solo games in \cref{tab:omni} using their native normalized scores and the mean across games. 
\red{For the main benchmarks, we run three trials per OpenHA task and five episodes per OmniGameArena game. For ViZDoom, we use the four-map protocol of~\cite{wang2025gametars,guo2026gamewam} and report mean reward over 50 episodes per map, with a 35\,Hz simulation tick rate.}

\paragraph{Comparisons.}
We include commercial VLMs, open-weight VLMs, and trained game policies based on human gameplay, RL, or both.
\red{The benchmark tables compare complete systems. The shared-backbone comparison in \cref{sec:exp:ablation} controls for differences in model initialization.}
Data scale refers to gameplay used for policy training; hour equivalents inferred from tokens or frames are approximate (\red{Supplementary \cref*{sec:supp:exp:accounting}}).

\subsection{\red{Benchmark Results}}
\label{sec:exp:bench}
\label{sec:exp:transfer}

\paragraph{\red{OpenHA: instruction following.}}
On OpenHA (\cref{tab:openha}), \textsc{Script2Play} reaches 69.6\% average success, exceeding the strongest prior trained policy, Game-TARS MoE-mini (64.5\%), by 5.1 percentage points.
\red{This is the highest average among the compared trained game policies, using 150.3\,h of synthetic demonstrations and behavior cloning.}

\red{We then test the same checkpoint on UE5 and Doom, neither of which appears in post-training, to measure transfer across visual appearance, dynamics, and task composition.}

\paragraph{OmniGameArena: transfer beyond the training worlds.}
\red{We evaluate only OmniGameArena's 3D instruction-following games to align with the task setting of our training data.}
\textsc{Script2Play} achieves a mean score of 0.640 (\cref{tab:omni}), the highest score among all compared agents.
It exceeds the best commercial VLM in this comparison, Gemini~3.1 Pro Preview (0.473), by 0.167, and the strongest human-gameplay policy, GameScaling-9B-S3 (0.361), by 0.279.
The latter uses 1{,}600\,h of human gameplay, compared with our 150.3\,h of synthetic demonstrations.
\red{This comparison demonstrates strong performance at a smaller training scale.}

\paragraph{ViZDoom: transfer of combat skills.}
\red{On ViZDoom (\cref{tab:vizdoom}), \textsc{Script2Play} scores 12.92 without ViZDoom trajectories or target-game RL, compared with 14.54 for the zero-shot Game-TARS MoE-mini, which uses a much larger model and substantially more gameplay data.}
GameWAM is trained directly on expert APPO trajectories from the four evaluation maps, whereas \textsc{Script2Play} transfers zero-shot.
Game-TARS is built on the substantially larger Seed1.5-VL mixture-of-experts backbone and uses explicit sparse thinking~\cite{wang2025gametars}; \textsc{Script2Play} is a 4B model that omits explicit reasoning traces to reduce inference latency.
\red{The Battle maps require the agent to search for health and ammunition as well as enemies, although the instruction only asks it to kill enemies.}
\red{These implicit subgoals and the need to find enemies outside the current view are difficult for our policy, which cannot track previously searched areas or supply locations from a single frame.}

\paragraph{Action sequences reveal how control transfers.}
\red{\Cref{fig:io} illustrates how the policy translates observations into actions across engines.}
\red{In OpenHA combat, the policy turns and advances toward the target before attacking. In GUI tasks, it combines mouse positioning with item-manipulation clicks.}
\red{In the held-out UE5 engine, ObstacleRun3D requires forward motion interleaved with timed jumps, while CueChase requires camera adjustment, approach, and interaction.}
\red{The same action-chunk representation supports these distinct control behaviors in Minecraft and UE5.}
\red{These successful episodes illustrate policy behavior; the benchmark scores measure its reliability.}
\red{Please refer to our \anon{supplementary materials}{\href{https://assettool-labs.github.io/script2play/}{project page}} for video demonstrations.}

\begin{figure}[t]
  \revisionfloat
  \centering
  \revisiongraphic[width=1.0\linewidth]{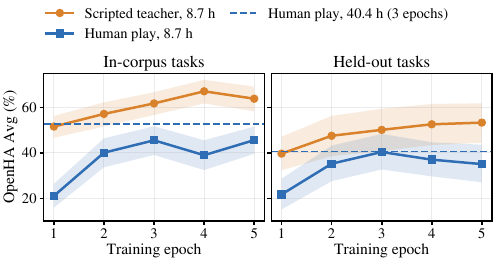}
  \caption{\textbf{Matched human versus scripted supervision.} OpenHA success by epoch for Qwen3.5-4B with matched per-task data budgets (8.7\,h per source). Bands show 95\% task-bootstrap confidence intervals. Dashed line: 40.4\,h human control trained for three epochs. Supplementary \cref*{tab:abl:matched} reports paired differences.}
  \label{fig:abl:matched}
\end{figure}

\Needspace{5\baselineskip}
\subsection{Effect of the Supervision Source}
\label{sec:exp:ablation}

\begin{revisionblock}
\paragraph{Shared-backbone comparison.}
We compare three policies initialized from Qwen2.5-VL-7B~\cite{bai2025qwen25vl}: OpenHA trained on human play, HiSTEVE trained with human-derived supervision and RL, and a separate \textsc{Script2Play} variant trained by behavior cloning on 26.9\,h of Minecraft teacher demonstrations, without \anon{in-house game engine}{Roblox} data.
This comparison fixes the backbone; data budgets and training procedures differ.
We compare OpenHA success rates for all three and evaluate OpenHA and our policy zero-shot on OmniGameArena; HiSTEVE weights are unavailable for the latter comparison.
Our policy achieves 45.9\% OpenHA success versus 31.5\% and 37.4\%, leading both baselines across all three task groups (\cref{tab:abl:qwen25}a), with similar zero-shot scores to OpenHA (0.028 versus 0.023; \cref{tab:abl:qwen25}b).
Thus, the in-domain advantage persists when the backbone is held fixed.

\paragraph{Matched human and scripted supervision.}
To isolate the supervision source, we train two Qwen3.5-4B policies on GameWAM's event-aligned human VPT clips~\cite{guo2026gamewam,baker2022vpt} or verified Minecraft teacher demonstrations.
We use 429 training tasks and hold out 136 tasks.
For each training task, we cap both sources at the smaller window count and sample whole clips, yielding approximately 8.7\,h of data per source.
Both use instructions sampled from OpenHA task descriptions, the same observation/action format and training recipe (\cref{sec:exp:data}), and five epochs (2{,}405 steps, batch size 384).
A larger-data human control uses all 40.4\,h available for the same training tasks, for three epochs (6{,}683 steps).

We evaluate each epoch on tasks from the OpenHA benchmark: 240 in-corpus tasks and all 136 held-out tasks.
The in-corpus tasks also appear in training, but all evaluations use environments excluded from both training sources.

\paragraph{Results.}
Scripted supervision leads at all five epochs on both task sets (\cref{fig:abl:matched}), with 9.7--18.3-point held-out gains.
Its one-epoch in-corpus success (51.6\%) exceeds the best matched-human checkpoint (45.6\%); from epoch two, it also beats the 40.4\,h control.
Gains are largest on Embodied and Combat; the larger human control leads on GUI.
Scripted supervision improves data efficiency and generalization.

\subsection{Which Skills Transfer?}
\label{sec:exp:skills}

We train seven Qwen3.5-4B policies on all nonempty subsets of Combat \& Shooting (C), Traversal \& Parkour (T), and Dodging \& Survival (D). We evaluate transfer on three OmniGameArena games, one per skill category: Monster Shoot (C), Obstacle Run3D (T), and Last Stand (D) (\cref{tab:skills}).
This ablation uses subsets of the \anon{in-house game engine}{Roblox} data, containing 20.5, 12.3, and 15.8\,h for C, T, and D, respectively.
All arms use the \textsc{Script2Play} training recipe for 3{,}000 steps, with no Minecraft data. Within each batch, the selected categories share 85\% equally, and web data supplies the remaining 15\%.
Each policy plays 10 episodes per game under the time-paused protocol.

Without the corresponding training category, Monster Shoot scores 0 and the other two games remain near Random.
Adding traversal data hurts Last Stand: policies keep moving forward and fall sooner. Adding combat data also halves the dodging-data share without the same drop.
Transfer therefore depends on both relevant skill coverage and the compatibility of combined skills.
\end{revisionblock}

\FloatBarrier 
\section{Conclusion}
\label{sec:conclusion}

\red{We presented a fully automated framework that converts gameplay skill specifications into games, task verifiers, and programmatic experts.}
\red{Behavior cloning on 150.3\,h of synthetic gameplay in Minecraft and \anon{an in-house game engine}{Roblox} yields \textsc{Script2Play}, a unified 4B VLA policy trained without human gameplay demonstrations or online RL.}
\red{A single checkpoint is evaluated across three benchmarks, leading the compared trained policies on OpenHA and achieving the highest mean score on the evaluated OmniGameArena games.}
\red{These results show that programmatic action supervision can turn pretrained visual knowledge into transferable game control.}
\red{Extending skill coverage to more complex interactions and longer-horizon tasks is a natural next step.}

\clearpage 
{
    \small
    \bibliographystyle{ieeenat_fullname}
    \bibliography{main}
}

\maketitlesupplementary
\appendix
\let\paragraph\supplementaryparagraph
\setcounter{figure}{0}
\setcounter{table}{0}
\setcounter{algorithm}{0}
\renewcommand{\thefigure}{S\arabic{figure}}
\renewcommand{\thetable}{S\arabic{table}}
\renewcommand{\thealgorithm}{S\arabic{algorithm}}
\def\theHfigure{S\arabic{figure}}
\def\theHtable{S\arabic{table}}
\def\theHalgorithm{S\arabic{algorithm}}
\section{Game Generation and Validation}
\label{sec:supp:generated}

\red{This section details the \anon{in-house game}{Roblox game} generation workflow of main-paper \cref*{sec:method:generation}, from skill specifications to verified demonstrations.}
Each specification describes the target skills, camera, dynamics, controls, and a task contract defining the instructed target, completion condition, and prohibited events.
The coding agent implements the game, its scene generator, a scripted teacher, and a task verifier. It also writes a Python implementation of the game rules, called the \emph{offline mirror}, for testing the teacher and verifier before engine validation.
\red{The scene generator varies layouts, appearances, and dynamics; instruction paraphrases vary the wording while preserving task meaning.}
Paired tasks share a scene but request different targets; each member is a task variant, or \emph{arm}.
\red{The audit checks that a teacher pursuing one variant's target does not pass the other variant's verifier.}

\paragraph{The full loop.}
\Cref{alg:pipeline} summarizes game generation, verification, demonstration collection, and behavior cloning.

\begin{algorithm}[ht]
  \caption{Game generation and policy training.}
  \label{alg:pipeline}
  \small
  \begin{algorithmic}
    \Require specifications $\mathcal{Z}$; repair budget $R=3$
    \State $\mathcal{G} \gets \emptyset$
    \For{each $z \in \mathcal{Z}$}
      \State $G \gets \textsc{Generate}(z)$; $r \gets 0$
      \Loop
        \State $\rho \gets \textsc{Audit}(G)$ \Comment{G1--G5}
        \State \textbf{if} $\rho$ passes \textbf{then} $\mathcal{G} \gets \mathcal{G} \cup \{G\}$; \textbf{break}
        \State \textbf{if} static or wrong-target checks fail \textbf{then break}
        \State \textbf{if} $r=R$ \textbf{then break}
        \State $\textsc{Retune}(G, \rho)$ \Comment{widen or narrow tolerances}
        \State $r \gets r+1$
      \EndLoop
    \EndFor
    \State $\mathcal{D} \gets \textsc{VerifiedRollouts}(\mathcal{G})$
    \State $\pi_\theta \gets \textsc{BC}(\mathcal{D})$
    \State \Return $\pi_\theta$
  \end{algorithmic}
\end{algorithm}

\paragraph{Specification fields.}
\Cref{tab:spec} describes the specification fields and their role in game generation.

\begin{table}[ht]
  \centering
  \caption{\textbf{Specification fields.} The specification guides game mechanics, task construction, and teacher behavior.}
  \label{tab:spec}
  \small
  \setlength{\tabcolsep}{3pt}
  \begin{tabular}{@{}>{\raggedright\arraybackslash}p{1.7cm}>{\raggedright\arraybackslash}p{6.2cm}@{}}
    \toprule
    Field & Examples \\
    \midrule
    Skill & a gameplay category from main-paper \cref*{fig:data3d:mosaic}, \eg traversal and parkour \\
    Camera & first person, third-person follow, top-down, side view \\
    Dynamics & ranges for speeds, tolerances, hold durations, gravity, gap widths \\
    Controls & WASD or arrows, Space, E/F, left button, mouse aim, with a plain-language description \\
    Contract & requested target (\eg a pad or gate); completion condition; prohibited events (\texttt{wrong\_target}, \texttt{fell}, \texttt{over\_budget}) \\
    \bottomrule
  \end{tabular}

\end{table}

\paragraph{Validation gates.}
\Cref{tab:gates} summarizes the audit checks.
\red{G2 and G3 test the teacher and verifier using the Python game rules. G4 then runs the teacher in the game engine to check whether these results carry over to engine execution and input timing. G5 checks the recorded images.}

\begin{table}[ht]
  \centering
  \caption{\textbf{Game-validation checks.} Games must pass all five gates before their demonstrations enter the training corpus.}
  \label{tab:gates}
  \small
  \setlength{\tabcolsep}{3pt}
  \begin{tabular}{@{}>{\raggedright\arraybackslash}p{1.75cm}>{\raggedright\arraybackslash}p{6.15cm}@{}}
    \toprule
    Check & Purpose \\
    \midrule
    G1: Static validity & Check that the code compiles and that scene, task, and wording use separate random streams. Paired tasks must share a scene while requesting different targets. \\
    G2: Solvability & Require at least 95\% teacher success in Python tests, both under nominal timing and with added latency. \\
    G3: Non-triviality & Check that idle, hold-forward, random-key, and wrong-target policies rarely succeed. These tests detect shortcuts and verifiers that ignore the instructed target. \\
    G4: Engine parity & Run the same teacher in the game engine. Its success rate must be within five percentage points of the Python tests in G2. \\
    G5: Visual integrity & Check that decoded recordings contain non-blank frames that vary during play. Inspect appearance variation across recordings. \\
    \bottomrule
  \end{tabular}

\end{table}

\red{A task can be easy for the teacher yet provide little useful supervision.}
\red{The audit therefore checks for goals already satisfied at spawn, tasks solved by waiting, and verifiers that accept the wrong target.}
\red{For sustained actions, the verifier distinguishes a continuous hold from time accumulated over separate attempts.}
These checks screen out tasks that can be completed without the intended behavior.

\paragraph{Repair and acceptance.}
For recoverable failures, the agent adjusts task tolerances and reruns the audit, for up to three repair rounds.
Wider tolerances can make a task solvable, while narrower tolerances can eliminate trivial solutions.
\red{We discard games that fail static-validity or wrong-target checks, or exhaust the repair budget without passing the audit.}
Accepted games proceed to demonstration collection.

\paragraph{Teacher decision loop.}
\Cref{alg:teacher} describes the teacher's latency-compensated decision loop (main-paper \cref*{sec:method:training}).
\red{At each control step, the teacher reads privileged state and accounts for previously issued inputs that have not yet taken effect.}
\red{It then forms a mechanic-specific plan $p_t$ and converts it to the keyboard-and-mouse action $a_t$ used by the student.}

\begin{algorithm}[ht]
  \caption{Latency-compensated teacher control.}
  \label{alg:teacher}
  \small
  \begin{algorithmic}
    \Require instruction $g$; state $s_t$; controls $c$; pending keys $K$ and mouse motion $M$ at control step $t$
    \State $\hat{s}_t \gets \textsc{Advance}(s_t, K, M)$ \Comment{latency compensation}
    \State $p_t \gets \pi^*_{\mathrm{mech}}(\hat{s}_t, g, c)$ \Comment{\eg lead, aim, fire}
    \State $a_t \gets \textsc{Chunks}(p_t)$ \Comment{key groups and mouse motion}
    \If{pending motion risks overshooting the target}
      \State Adjust $a_t$ to release held keys before overshoot
    \EndIf
    \State \Return $a_t$ aligned with the current observation $o_t$
  \end{algorithmic}
\end{algorithm}

\paragraph{Demonstration collection.}
\red{The teacher acts through the game's normal input interface, while the recorder aligns rendered observations with instructions and action chunks.}
The verifier checks completion and prohibited events throughout the episode.
Only successful, verified rollouts enter the training corpus.
Behavior cloning then trains the student on the recorded instructions, images, and actions; privileged state remains available only to the teacher and verifier.

\paragraph{Optional feedback for generation.}
\red{Failure analysis can guide new specifications by grouping episodes by skill, camera, or control type. For example, repeated falls or hazard collisions can motivate simpler variants of a task. Any resulting game must pass the same validation gates before contributing demonstrations.}

\section{Coding Agent, Tools, and an Example Game}
\label{sec:supp:agent}

\paragraph{Coding agent.}
\red{Cursor Agent with Claude Opus 5.5 wrote the \anon{in-house}{Roblox} games shown in main-paper \cref*{fig:data3d:mosaic}, working on one game at a time with access to the game repository and run directories.}
For each game, the agent writes four components and registers the task templates.
The \emph{scenario module}, written in \red{\anon{the engine's server-side scripting language}{server-side Luau}}, declares the templates and constructs an episode from its seed.
The \emph{game module}, written in \red{\anon{the engine's client-side scripting language}{client-side Luau}}, implements the mechanics in a custom fixed-step simulation at 60\,Hz, \red{without using \anon{the built-in}{Roblox's} physics engine}.
The \emph{teacher} is a Python controller that reads published engine state and returns a 200\,ms action chunk containing four 50\,ms key groups and mouse motion.
The \emph{offline mirror} tests the teacher in closed loop using a Python implementation of the game rules.
\red{The agent runs the audit, recording, and export commands, then inspects recorded frames.}
\red{Researchers supplied the specifications and shared asset packs. During development, they also spot-checked each new game's code and a short video; the agent implemented the fixes.}

\paragraph{Tools.}
\red{The agent uses file-reading and editing tools, a shell, and an image viewer.}
\red{\anon{Command-line tools assemble}{Rojo assembles} each \anon{environment}{Roblox environment} from generated code, without manual editing in \anon{a graphical editor}{Roblox Studio}.}
\red{The generated code calls a shared runtime library for environment editing, terrain, asset placement, and scripting. Audit checks run as command-line tools (\cref{tab:agent:api}).}

\begin{table}[ht]
  \centering
  \caption{\textbf{Tools available to the coding agent.} Shared runtime functions construct episodes and expose state; command-line tools validate and record the games.}
  \label{tab:agent:api}
  \small
  \setlength{\tabcolsep}{3pt}
  \begin{tabular}{@{}>{\raggedright\arraybackslash}p{1.6cm}>{\raggedright\arraybackslash}p{6.3cm}@{}}
    \toprule
    Layer & Calls \\
    \midrule
    Episode & \texttt{params} samples template parameters; \texttt{choose} selects the arm's target; \texttt{say} generates the instruction and paraphrases; \texttt{contract} specifies the target, effect, and constraints; \texttt{part} places geometry. Scene, appearance, task, and text use separate random streams: paired arms share a scene, and appearance changes preserve its layout. \\
    World & \texttt{world} surrounds the arena with one of seven themes (terrain, \red{\anon{asset-library}{Toolbox} props}, sky, lighting), assigns \red{\anon{character}{Roblox} avatars} to the player and NPCs, and samples camera settings \\
    Game & \texttt{fixedStep} (simulation); \texttt{pressed}, \texttt{key}, \texttt{mouseLook} (input); \texttt{part}, \texttt{model}, \texttt{figure}, \texttt{chaseCamera} (scene); \texttt{publish} (the state the teacher reads); \texttt{complete}, \texttt{violate} (the verifier) \\
    Checks & \red{\anon{Static code analyzer}{Luau analyzer}} (G1); mirror tests (G2--G3); teacher rollouts in the headless engine, recorded at 20\,fps (G4); export to 200\,ms windows; recorded-frame checks (G5) \\
    \bottomrule
  \end{tabular}

\end{table}

\begin{figure*}[t]
  \centering
  \setlength{\tabcolsep}{1.5pt}
  \renewcommand{\arraystretch}{0.6}
  \begin{tabular}{@{}*{4}{>{\centering\arraybackslash}p{0.243\linewidth}}@{}}
    \multicolumn{4}{@{}l}{\small\textbf{Arm 0:} ``Wait for the laser, jump the gap and stop on the red marker.''} \\
    \includegraphics[width=\linewidth]{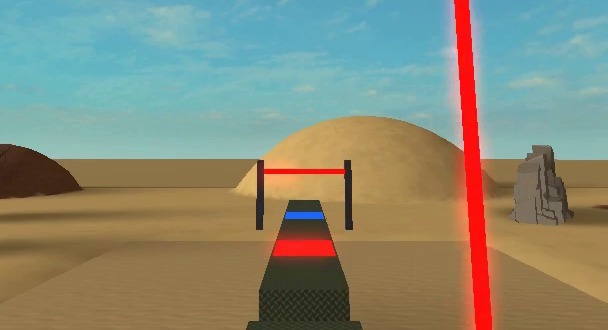}\newline{\scriptsize 0.85\,s, \texttt{W}\par} &
    \includegraphics[width=\linewidth]{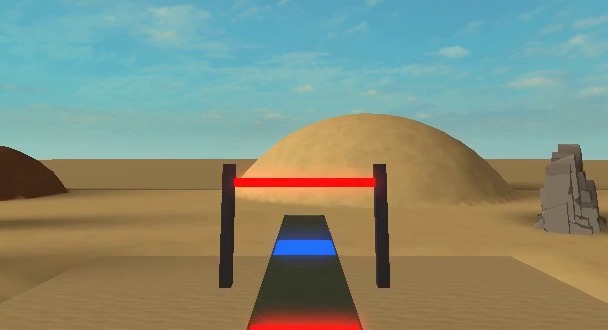}\newline{\scriptsize 2.60\,s, \texttt{W+Space}\par} &
    \includegraphics[width=\linewidth]{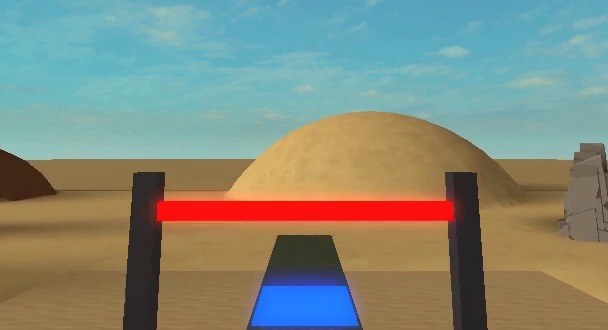}\newline{\scriptsize 3.75\,s, \texttt{W}\par} &
    \includegraphics[width=\linewidth]{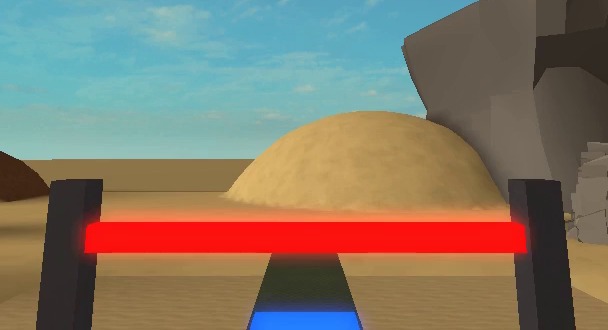}\newline{\scriptsize 5.20\,s, no key\par} \\[2pt]
    \multicolumn{4}{@{}l}{\small\textbf{Arm 1:} ``Wait for the laser, jump the gap, crouch under the low laser and stop on the blue marker.''} \\
    \includegraphics[width=\linewidth]{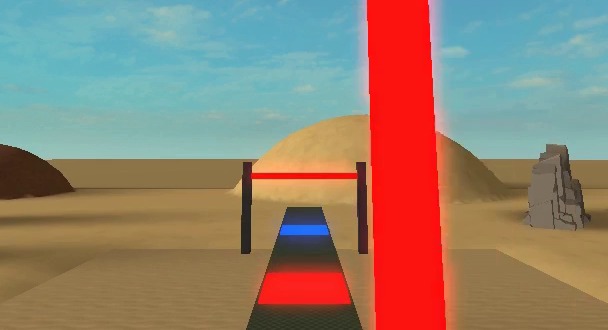}\newline{\scriptsize 1.90\,s, no key\par} &
    \includegraphics[width=\linewidth]{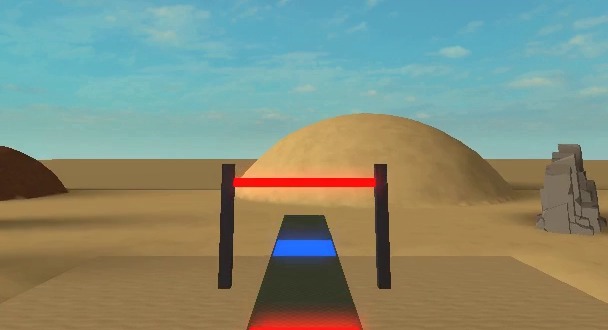}\newline{\scriptsize 2.80\,s, \texttt{W+Space}\par} &
    \includegraphics[width=\linewidth]{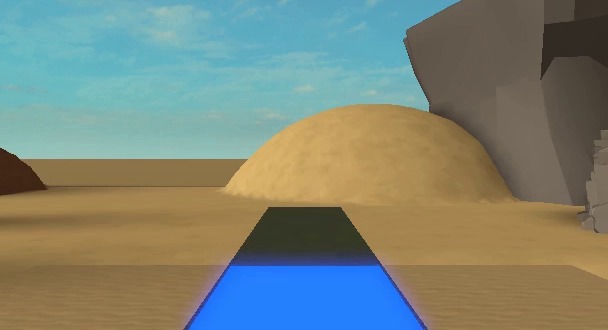}\newline{\scriptsize 5.00\,s, \texttt{W+C}\par} &
    \includegraphics[width=\linewidth]{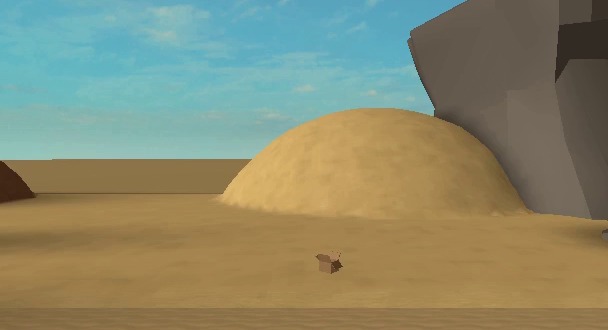}\newline{\scriptsize 7.15\,s, no key\par} \\[2pt]
\end{tabular}

  \caption{\red{\textbf{The generated Laser Maintenance Bridge.} Paired tasks use the same desert scene but specify different destinations. Arm~0 passes the sweeping laser, jumps the gap, and stops at the red marker before the low laser. Arm~1 continues under the low laser to the blue marker, which is beneath the player in the final frame. Labels show elapsed time and held keys.}}
  \label{fig:example:game}
\end{figure*}

\paragraph{Prompt.}
\red{The prompt names the game and components to read first. A repository guide limits execution to eight engine workers, permits one recording at a time, and forbids game-file edits during recording. The example below uses generic component names.}

\noindent\fbox{\begin{minipage}{\dimexpr\linewidth-2\fboxsep-2\fboxrule\relax}\small\raggedright
Add a game for \texttt{<specification>} following the repository's game-creation guide.
First read the shared runtime, task-specification and episode-context components, and the closest existing game's scenario, mechanics, teacher, and Python tests.
Implement and register the new components. Reach at least 95\% teacher success for both task variants in the Python tests, then run a 16-episode engine check. Show frames from a successful episode and, if available, a failed episode.
\end{minipage}}

\paragraph{Example: Laser Maintenance Bridge.}
\Cref{tab:example:spec} specifies a traversal-and-parkour game; \cref{fig:example:game} shows paired instructions in one generated scene.
\red{The teacher faces the camera down the bridge and predicts the player's position when the next action takes effect, accounting for pending inputs.}
\red{For the \texttt{route} template, it waits for the sweeping laser, jumps the gap while holding \texttt{W}, and stops at the requested marker. Reaching the second marker also requires crouching under the low laser with \texttt{C}.}
It estimates input latency online from the distance traveled after issuing \texttt{W}.
In the Python tests, the teacher passed all 120 episodes (40 per template) under nominal timing and 119 of 120 with three extra simulation frames of latency and 12--24-frame decision intervals; the single failure was a laser contact.
The engine check passed all 48 episodes (16 per template, covering both arms).
Of the 990 recorded episodes (165 counterfactual pairs per template), 988 passed, yielding 1.6\,h of demonstrations recorded at 20\,fps.

\begin{table}[ht]
  \centering
  \caption{\textbf{Laser Maintenance Bridge specification.} The five fields in \cref{tab:spec} and the resulting task templates. Distances and speeds use simulation units and units/s, respectively.}
  \label{tab:example:spec}
  \small
  \setlength{\tabcolsep}{3pt}
  \begin{tabular}{@{}>{\raggedright\arraybackslash}p{1.6cm}>{\raggedright\arraybackslash}p{6.3cm}@{}}
    \toprule
    Field & Laser Maintenance Bridge \\
    \midrule
    Skill & Traversal and parkour: timed jumps, waiting for moving hazards, and controlled movement \\
    Camera & First-person view on a bridge of width 3 above a drop; initial view direction within $\pm 20^\circ$ of the bridge's axis \\
    Dynamics & Gap width: 2.4--3.0 (running-jump range: 5). Sweeping-laser speed: 5--8; sweep half-width: 5--7. Low-laser height: 2.9--3.4; player height: 4.8 standing, 2.4 crouched. Hazard spacing: 3--5. Movement speed: 7 walking, 3.5 crouching. \\
    Controls & ``The mouse turns your view and W, A, S and D walk. Space jumps. Hold C to crouch (and walk slower).'' \\
    Contract & Instructed variable: the first or the second marker (red and blue, colors shuffled); effect: reach it and stand still (0.3\,s), then 0.8\,s without a violation; violations: \texttt{fell}, \texttt{laser}, \texttt{too\_far}; 14\,s budget \\
    Templates & \texttt{gap\_jump}: cross one gap to the first marker or two to the second; \texttt{laser\_wait}: pass one sweeping laser or two; \texttt{route}: pass a sweeping laser and a gap to the first marker, then crouch under a low laser to reach the second. Each instruction has two paraphrases. \\
    \bottomrule
  \end{tabular}

\end{table}

\section{Experimental Data Accounting}
\label{sec:supp:exp:accounting}

\paragraph{Our training mixture.}
The main \textsc{Script2Play} policy is initialized from Qwen3.5-4B~\cite{qwen2026qwen35} and trained for 8{,}000 steps on 150.3\,h of synthetic demonstrations: 26.9\,h from Minecraft and 123.4\,h from \red{\anon{the in-house game engine}{Roblox}}.
This checkpoint provides the main results on OpenHA, OmniGameArena, and ViZDoom in main-paper \cref*{tab:openha,tab:omni,tab:vizdoom}.
The separate Qwen2.5-VL-7B variant~\cite{bai2025qwen25vl} uses only the Minecraft component.
These gameplay durations exclude the non-gameplay web data, which occupies 15\% of each training batch.
Our policies are trained by behavior cloning, without RL or DAgger.

\paragraph{Baseline scales.}
Main-paper \cref*{tab:openha,tab:vizdoom,tab:abl:qwen25} mark approximate hour equivalents with $^\dagger$.
For frame counts, we use 20\,fps, or 72{,}000 frames per hour. For token counts, the rough conversion assumes 350 tokens per frame, or approximately 25M tokens per hour. This token-to-time ratio is an accounting assumption, not a measured property shared by the models.

ROCKET-1's 1.6B frames correspond to approximately 22{,}000\,h~\cite{cai2025rocket1}. STEVE-1's approximately 70{,}750\,h combines 70{,}000\,h of inherited VPT pretraining with 54M fine-tuning frames, or 750\,h at 20\,fps~\cite{baker2022vpt,lifshitz2023steve1}.
For JARVIS-VLA, treating 7.4M gameplay frames and 6.4M GUI entries as frame equivalents gives approximately 192\,h~\cite{li2025jarvisvla}; GUI entries are not measurements of continuous gameplay time. HiSTEVE also uses VPT, JARVIS-VLA, and additional scripted trajectories~\cite{tang2026histeve}, so JARVIS-VLA's counts do not specify its full corpus duration.
OpenHA's 3.62B consumed tokens, reported in GameWAM's comparison~\cite{guo2026gamewam}, give approximately 145\,h under the token conversion. Game-TARS separately reports approximately 20{,}000\,h of pretraining gameplay and 2{,}000\,h for post-training~\cite{wang2025gametars}.

\paragraph{Corpus duration and training exposure.}
GameWAM reports 200{,}000 event-centered VPT clips of 96 frames, alongside regular VPT and scripted GUI data~\cite{guo2026gamewam}. The event clips contain 19.2M retained frames, or 266.7\,h at 20\,fps; filtering and overlap prevent treating this as unique source-video duration. Its 2.79B consumed tokens and 7{,}708{,}800 training-sample presentations ($21{,}900\times352$) instead describe exposure during optimization. Neither count establishes the duration of a unique ViZDoom corpus.
\red{Corpus frames, GUI entries, consumed tokens, and repeated training samples therefore measure different quantities. The hour equivalents provide approximate scale context, not a uniform measure of unique gameplay or a controlled data-efficiency ratio.}

\begin{revisionblock}
\section{Matched Human and Scripted Supervision}
\label{sec:supp:exp:matched}

\begin{table}[htbp]
  \revisionfloat
  \centering
  \caption{\textbf{Smaller matched budgets} (OpenHA in-corpus Avg, \%). Each row trains both sources on the given fraction of every training task's budget and scores the in-corpus tasks present in both training sets. Task sets vary across rows; source comparisons are paired within each row. $\Delta$ is scripted minus human, with a paired 95\% confidence interval.}
  \label{tab:supp:matched:budgets}
  \footnotesize
  \setlength{\tabcolsep}{0pt}
  \newcommand{\oci}[2]{\makecell[c]{#1\\[-2.5pt]{\scriptsize[#2]}}}
  \begin{tabular*}{\linewidth}{@{\extracolsep{\fill}}lccccc@{}}
    \toprule
    Budget & \makecell{Data\\(human / teacher)} & Tasks & Human & Scripted & $\Delta$ \\
    \midrule
    1/16 & 0.53\,h / 0.53\,h & 188 & 7.2 & 18.2 & \oci{+11.0}{7.3, 14.8} \\
    1/8 & 1.07\,h / 1.10\,h & 211 & 9.8 & 20.3 & \oci{+10.5}{4.1, 17.0} \\
    1/4 & 2.18\,h / 2.17\,h & 225 & 14.3 & 41.8 & \oci{+27.5}{20.6, 34.4} \\
    1/2 & 4.40\,h / 4.36\,h & 239 & 30.0 & 50.4 & \oci{+20.4}{14.8, 26.2} \\
    \bottomrule
  \end{tabular*}

\end{table}

\begin{table*}[t]
  \revisionfloat
  \centering
  \caption{\textbf{Matched human versus scripted supervision on OpenHA} (success rate, \%). The two 8.7\,h policies use Qwen3.5-4B, the same training-task split, and matched per-task data budgets. In-corpus: 240 tasks that also appear in training; held-out: 136 tasks excluded from training. All evaluations use environments excluded from both training sources. Avg is the mean of the three groups, with 95\% task-bootstrap confidence intervals. $\Delta$ is scripted minus human Avg at the same epoch, with paired intervals. The last row is a larger-data control using all 40.4\,h of human play for the training tasks. See Appendix~\ref{sec:supp:exp:matched} for the protocol.}
  \label{tab:abl:matched}
  \footnotesize
  \setlength{\tabcolsep}{1.5pt}
  \renewcommand{\arraystretch}{1.06}
  \newcommand{\oci}[2]{\makecell[c]{#1\\[-2.5pt]{\scriptsize[#2]}}}
  \begin{tabular*}{\textwidth}{@{\extracolsep{\fill}}lllcccccccccc@{}}
    \toprule
    & & & \multicolumn{5}{c}{In-corpus (240 tasks)} & \multicolumn{5}{c}{Held-out (136 tasks)} \\
    \cmidrule(lr){4-8}\cmidrule(l){9-13}
    Epoch & Source & Data & Emb. & GUI & Combat & Avg & $\Delta$ & Emb. & GUI & Combat & Avg & $\Delta$ \\
    \midrule
    1 & Scripted & 8.7\,h & 64.0 & 8.0 & 82.8 & \oci{51.6}{46.6, 56.3} & \oci{+30.5}{24.5, 36.4} & 62.2 & 6.8 & 50.0 & \oci{39.7}{32.3, 47.1} & \oci{+18.0}{11.6, 24.0} \\
     & Human & 8.7\,h & 33.3 & 4.0 & 25.9 & \oci{21.1}{15.9, 26.4} &  & 36.7 & 3.4 & 25.0 & \oci{21.7}{15.1, 28.7} &  \\
    \addlinespace[2pt]
    2 & Scripted & 8.7\,h & 61.7 & 23.5 & 86.2 & \oci{57.1}{51.6, 62.2} & \oci{+17.1}{10.5, 23.8} & 63.3 & 21.6 & 57.7 & \oci{47.5}{38.5, 56.3} & \oci{+12.3}{4.7, 19.9} \\
     & Human & 8.7\,h & 44.6 & 20.5 & 55.2 & \oci{40.1}{33.7, 46.4} &  & 49.5 & 21.6 & 34.6 & \oci{35.2}{27.6, 43.2} &  \\
    \addlinespace[2pt]
    3 & Scripted & 8.7\,h & 65.3 & 35.5 & 84.5 & \oci{61.8}{56.7, 66.7} & \oci{+16.2}{10.0, 22.4} & 64.8 & 31.8 & 53.8 & \oci{50.1}{40.8, 59.4} & \oci{+9.7}{2.1, 17.9} \\
     & Human & 8.7\,h & 43.2 & 33.0 & 60.3 & \oci{45.5}{39.1, 51.7} &  & 52.6 & 37.8 & 30.8 & \oci{40.4}{32.7, 48.4} &  \\
    \addlinespace[2pt]
    4 & Scripted & 8.7\,h & 67.1 & 44.5 & 89.7 & \oci{67.1}{61.8, 72.1} & \oci{+28.1}{21.1, 35.2} & 66.8 & 35.1 & 55.8 & \oci{52.6}{43.4, 61.5} & \oci{+15.6}{6.0, 25.3} \\
     & Human & 8.7\,h & 41.4 & 25.5 & 50.0 & \oci{39.0}{32.5, 45.5} &  & 46.9 & 25.7 & 38.5 & \oci{37.0}{29.7, 44.7} &  \\
    \addlinespace[2pt]
    5 & Scripted & 8.7\,h & 65.3 & 40.0 & 86.2 & \oci{63.8}{58.3, 69.1} & \oci{+18.2}{12.4, 24.2} & 67.3 & 31.1 & 61.5 & \oci{53.3}{44.5, 61.8} & \oci{+18.3}{11.1, 25.5} \\
     & Human & 8.7\,h & 49.5 & 23.5 & 63.8 & \oci{45.6}{39.6, 51.6} &  & 46.9 & 23.6 & 34.6 & \oci{35.1}{27.1, 43.6} &  \\
    \midrule
    3 & Human & 40.4\,h & 46.4 & 51.5 & 60.3 & \oci{52.7}{46.6, 58.7} &  & 49.0 & 45.9 & 26.9 & \oci{40.6}{32.6, 49.5} &  \\
    \bottomrule
  \end{tabular*}

\end{table*}

\paragraph{Data.}
Human demonstrations are drawn from the event subset of GameWAM's Minecraft data~\cite{guo2026gamewam}: 96-frame clips of VPT contractor play~\cite{baker2022vpt}, labeled by the OpenHA task event they contain.
We exclude GameWAM's scripted GUI episodes, unlabeled free-play trajectories, and validation split, leaving only human play.
Teacher demonstrations come from the verified Minecraft data described in main-paper \cref*{sec:exp:data}.
Both sources use the observation and action format of our main model.

\paragraph{Tasks.}
The experiment uses 565 OpenHA tasks covered by both sources, after excluding four mining tasks whose OpenHA names do not match event labels.
We hold out tasks within each task type: 30\% of the mining and the mob-combat tasks (always including \texttt{kill\_entity:zombie}), 20\% of the crafting tasks, and five interaction and five smelting tasks.
This leaves 429 training tasks (111 Embodied, 288 GUI, 30 Combat) and 136 held-out tasks (49 Embodied, 74 GUI, 13 Combat).

\paragraph{Matching and training.}
For each training task, we set the budget to the smaller of the two sources' window counts, then sample whole clips in a fixed shuffled order up to that budget.
The resulting sets contain 156{,}819 human and 156{,}954 teacher windows, corresponding to 8.71\,h and 8.72\,h at 200\,ms per window. Whole-clip sampling leaves a small difference in the final counts.
For each window, we sample an instruction from OpenHA's descriptions of the task using a fixed random seed.
Both policies start from Qwen3.5-4B and follow the main-paper training recipe for five epochs (2{,}405 steps, batch size 384), with 85\% gameplay and 15\% web data in each batch. We save a checkpoint after each epoch (every 481 steps).
The 40.4\,h human policy uses every human window of the training tasks (727{,}106 windows) and is trained for three epochs (6{,}683 steps).

\paragraph{Evaluation.}
All evaluations use OpenHA benchmark tasks in environments excluded from both training sources. In-corpus tasks also appear in training; held-out tasks do not.
The initial in-corpus pool contains 241 tasks: all 141 Embodied and Combat training tasks, plus 100 GUI tasks sampled in proportion to task type. After retaining tasks with data in both sampled training sets, we evaluate 240 tasks with two trials each.
We also evaluate all 136 held-out tasks, with four trials per Embodied or Combat task and two per GUI task.
We first average trial success within each task, then average tasks within each group. Avg is the unweighted mean of the three group scores.
The 95\% confidence intervals use 10{,}000 bootstrap resamples of tasks within groups. For the scripted-minus-human difference $\Delta$, each resample uses the same tasks for both policies. \Cref{tab:abl:matched} reports all five epochs and the larger human-data control.

\paragraph{Smaller budgets.}
\Cref{tab:supp:matched:budgets} compares the two sources at nested fractions of each task's budget.
Keeping clips whole leaves some tasks without data at small budgets, and the missing tasks differ between sources. At each budget, we therefore score only in-corpus tasks present in both training sets.
Both policies train for 150 steps at the 1/16 budget and for three epochs at each larger budget.
The teacher-trained policy leads at every evaluated budget, supporting the benefit of scripted supervision when data are limited.

\end{revisionblock}

\FloatBarrier 
\section{\red{Detailed Results and Skill Coverage}}
\label{sec:supp:visualizations}

\red{Main-paper \cref*{fig:data3d:mosaic} shows skill coverage and training frames; \cref*{fig:diversity,fig:io} show environment diversity and closed-loop policy actions. \Cref{tab:abl:matched} reports matched-supervision results at each epoch.}

\begin{revisionblock}
\paragraph{Skill-category composition.}
The ablation in main-paper \cref*{sec:exp:skills} uses three subsets of \anon{the in-house game engine}{Roblox} data. Combat \& Shooting (C) includes Monster arena, Arena battle, Maze battle, Line defense, and Center defense (20.5\,h). Traversal \& Parkour (T) includes Sky course, Auto runner, Windshaft glider, Laser bridge, and Platform walkway (12.3\,h). Dodging \& Survival (D) includes Falling tiles, Deck watch, and Blocking arena (15.8\,h).

\paragraph{Training and evaluation.}
The seven policies cover all nonempty combinations of C, T, and D. Each starts from Qwen3.5-4B and trains for 3{,}000 steps, without Minecraft data. Selected categories share 85\% of each batch equally; web data supplies the remaining 15\%. We evaluate each policy on three OmniGameArena games under the time-paused protocol: Monster Shoot for combat, Obstacle Run3D for traversal, and Last Stand for dodging and survival. Main-paper \cref*{tab:skills} reports the scores.
\end{revisionblock}

\FloatBarrier

\end{document}